\documentclass[letterpaper]{article} 
\usepackage{aaai23}  
\usepackage{times}  
\usepackage{helvet}  
\usepackage{courier}  
\usepackage[hyphens]{url}  
\usepackage{graphicx} 
\usepackage{natbib}  
\usepackage{caption} 
\usepackage{algorithm}
\usepackage{algorithmic}
\nocopyright 
\usepackage{newfloat}
\usepackage{listings}
\DeclareCaptionStyle{ruled}{labelfont=normalfont,labelsep=colon,strut=off} 
\floatstyle{ruled}
\newfloat{listing}{tb}{lst}{}
\floatname{listing}{Listing}
\usepackage{xcolor}

\usepackage{tabu}
\usepackage{multirow}
\usepackage{amsmath}

\newcommand{\para}[1]{{\bf \noindent #1 \hspace{0pt}}}
\usepackage{enumitem}

\usepackage{graphicx}
\usepackage{textcomp}
\usepackage{xcolor}
\usepackage{enumitem}
\usepackage{soul}
\usepackage{subfigure}

\usepackage{ntheorem}
\makeatletter
\renewtheoremstyle{plain}
  {\item{\theorem@headerfont ##1\ ##2\theorem@separator}~}
  {\item{\theorem@headerfont ##1\ ##2\ (##3)\theorem@separator}~}
\makeatother

\usepackage{amssymb}

{\theoremheaderfont{\upshape\bfseries}
 \theorembodyfont{\normalfont\slshape}
 \newtheorem{definition}{Definition}
 \newtheorem{mtheorem}{Theorem}
\newtheorem{mlemma}{Lemma}
 }

\begin{document}

\title{PateGail: A Privacy-Preserving Mobility Trajectory Generator\\ with Imitation Learning}

\author{
    Huandong Wang\textsuperscript{\rm 1},
    Changzheng Gao\textsuperscript{\rm 1},
    Yuchen Wu\textsuperscript{\rm 2},
    Depeng Jin\textsuperscript{\rm 1},
    Lina Yao\textsuperscript{\rm 3},
    Yong Li\textsuperscript{\rm 1}
}
\affiliations{
    \textsuperscript{\rm 1}Beijing National Research Center for Information Science and Technology (BNRist), \\Department of Electronic Engineering, Tsinghua University, China\\
    \textsuperscript{\rm 2}Carnegie Mellon University, USA\\
    \textsuperscript{\rm 3}CSIRO's Data61 and University of New South Wales, USA\\
    \{wanghuandong,liyong07\}@tsinghua.edu.cn
}

\maketitle

\begin{abstract}
Generating human mobility trajectories is of great importance to solve the lack of large-scale trajectory data in numerous applications, which is caused by privacy concerns. However, existing mobility trajectory generation methods still require real-world human trajectories centrally collected as the training data, where there exists an inescapable risk of privacy leakage. To overcome this limitation, in this paper, we propose PateGail, a privacy-preserving imitation learning model to generate mobility trajectories, which utilizes the powerful generative adversary imitation learning model to simulate the decision-making process of humans. Further, in order to protect user privacy, we train this model collectively based on decentralized mobility data stored in user devices, where personal discriminators are trained locally to distinguish and reward the real and generated human trajectories. In the training process, only the generated trajectories and their rewards obtained based on personal discriminators are shared between the server and devices, whose privacy is further preserved by our proposed perturbation mechanisms with theoretical proof to satisfy differential privacy. Further, to better model the human decision-making process, we propose a novel aggregation mechanism of the rewards obtained from personal discriminators. We theoretically prove that under the reward obtained based on the aggregation mechanism, our proposed model maximizes the lower bound of the discounted total rewards of users. Extensive experiments show that the trajectories generated by our model are able to resemble real-world trajectories in terms of five key statistical metrics, outperforming state-of-the-art algorithms by over 48.03\%. Furthermore, we demonstrate that the synthetic trajectories are able to efficiently support practical applications, including mobility prediction and location recommendation.
\end{abstract}


\maketitle

\section{Introduction}

Human mobility trajectory data is instrumental for a large number of applications. For example, for the mobile Internet service providers, based on mobility trajectories, the movement and communication process of mobile users can be simulated to implement a reliable and efficient performance evaluation of the mobile networks~\cite{Hess2016Data}. For the government, mobility trajectories can characterize the travel demand of the population and the traffic condition of the city, and thus provide important guidance to the transportation system planning~\cite{feng2019identification}. However, the utilization of real-world mobility trajectories leads to a growing privacy concern, since sensitive information of users can be leaked from their trajectories, e.g., which places they have visited and who they have met. Thus, it is hard to obtain a large-scale human mobility trajectory dataset to support numerous downstream applications. Under these circumstances, simulating human mobility behavior to produce realistic and high-quality mobility trajectory data becomes an important task for downstream applications, and has drawn much attention from both academia and industry.

\begin{figure} [t!]
\begin{center}
\includegraphics*[width=8.5cm]{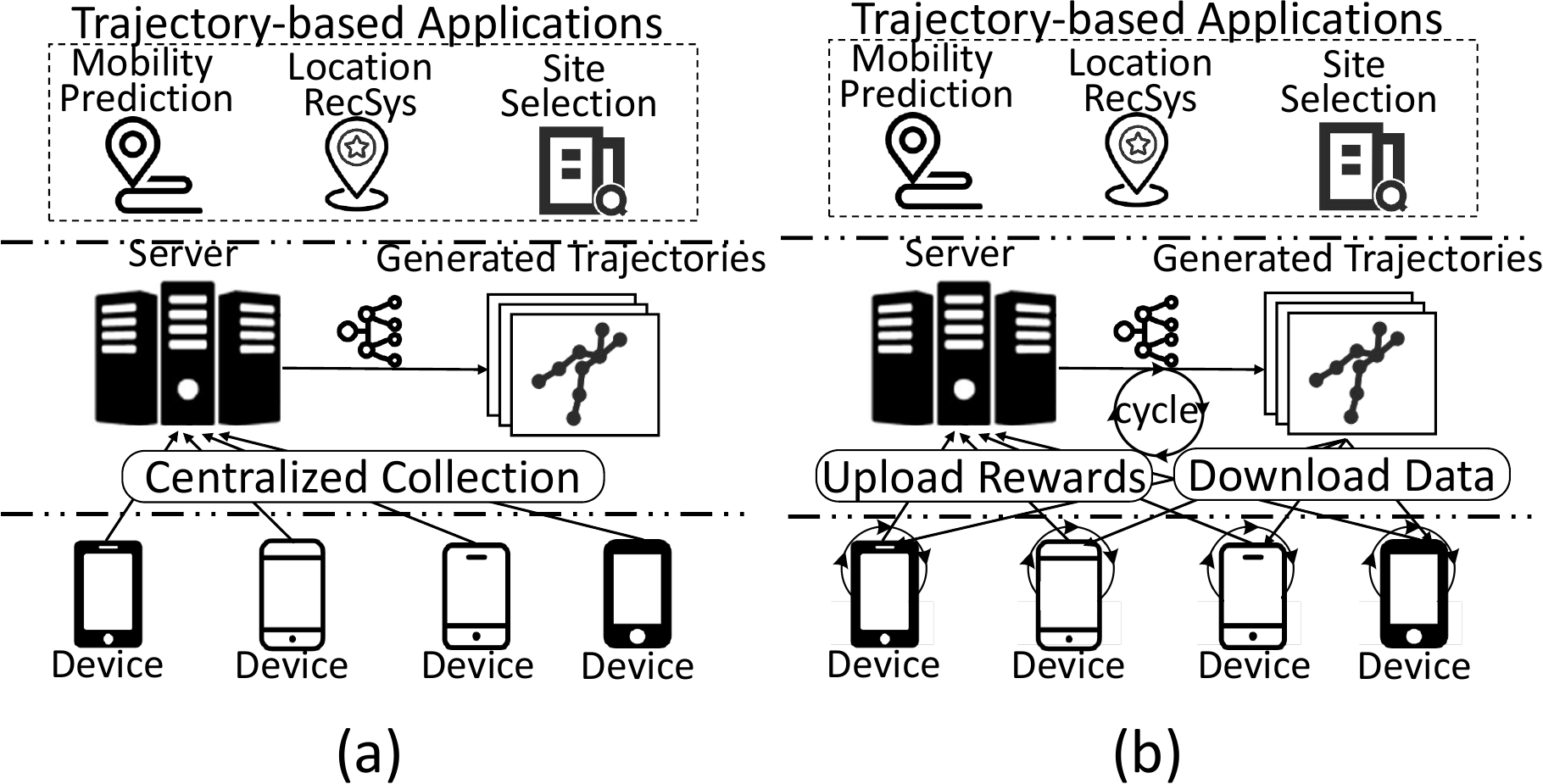}
\end{center}
\caption{Illustration of (a) existing trajectory generator and our
proposed (b) federated trajectory generator.}\label{fig:fig1}
\end{figure}

Numerous existing approaches have been proposed to generate mobility trajectories by utilizing powerful deep learning techniques including variational autoencoder (VAE)~\cite{huang2019variational}, and generative adversarial network (GAN)~\cite{feng2018deepmove,ouyang2018non,kulkarni2018generative,Liu2018trajGANsU}, etc. However, as shown in Figure~\ref{fig:fig1}(a), these methods still require a number of real-world human trajectories centrally collected as the training data, where there exists the risk of privacy leakage.
The rising paradigm of federated learning has provided a promising solution to this problem, which is a distributed machine learning framework with the goal of training machine learning models based on data distributed across multiple devices and protecting users' privacy at the same time. Federated learning has shown success in a number of practical applications, including personalized recommendation~\cite{chen2018federated}, keyboard prediction~\cite{hard2018federated}, etc. 
Thus, we seek to train the mobility trajectory generator in the manner of federated learning. As shown in Figure~\ref{fig:fig1}(b), in the federated mobility trajectory generation system, each user device keeps the private mobility trajectory data belonging to its owner (user).
Only aggregated intermediate results proceeded by privacy protection mechanisms are shared between devices, while this system does not take any piece of the user trajectory data away from the device.
In this way, we can train the mobility trajectory generator without privacy leakage.

However, training an efficient trajectory generator based on federated learning is not an easy task with the following challenges. 
First, mobility trajectories are with high dimensions and complicated interactions with both spatial venues and timestamps. How to develop a trajectory generator that accurately models human mobility behavior is the first challenge. 
Second, users' privacy is still possible to be leaked from the transmitted intermediate results in the training process, while most existing solutions do not provide privacy-preserving guarantees of the training process~\cite{FengYXYWL20, ouyang2018non}. How to provide the privacy-preserving guarantees of the training process is the second challenge.

To overcome these challenges, in this paper, we propose PateGail, a privacy-preserving imitation learning based mobility trajectory generator. Specifically, this model utilizes the powerful technique of generative adversary imitation learning (GAIL) to extract the hidden human movement decision process correlated with both spatial venues and timestamps, and thus is able to produce plausible mobility trajectories with preserved utility. 
Further, in order to provide privacy-preserving guarantees,  we locally train a separate personal discriminator on each user device to distinguish and reward the generated and real-world decision-making sequences of human mobility, and then only share the generated trajectories and the rewards of the generated trajectories obtained based on personal discriminators between the devices and the server in the training process. Further, we propose a perturbation mechanism to prevent privacy leakage from the rewards of personal discriminators, which is theoretically proven to satisfy the differential privacy criterion.
Finally, we propose a novel aggregation mechanism based on the mean and variance of reward obtained from different personal discriminators, which is able to model the dynamics of reward function across users. Furthermore, we theoretically prove that under the reward obtained based on our proposed aggregation mechanism, our proposed model maximizes the lower bound of the discounted total rewards of users. Our contributions can be summarized as follows:
\begin{itemize}
    \item We propose a powerful mobile trajectory generator based on GAIL and federated learning, which is able to extract the hidden human decision process to generate plausible mobile trajectories and preserve user privacy with differential privacy guarantees at the same time.
    \item We propose a novel reward aggregation mechanism of reward obtained from personal discriminators of different users, which is able to model the dynamics of reward function across users. Furthermore, we theoretically prove that under our proposed reward aggregation mechanism, the obtained model maximizes the lower bound of the discounted total rewards of users.
    \item Extensive experiments show that the synthetic trajectories of our proposed model are able to preserve the statistical properties of the original dataset, and are able to efficiently support downstream applications by augmenting their training data.
    We release the code of our proposed algorithm as well as the datasets to better reproduce the experimental results\footnote{\url{https://github.com/tsinghua-fib-lab/PateGail}}.
\end{itemize}

\section{Mathematical Model and System Overview}

\begin{figure} [t!]
\begin{center}
\includegraphics*[width=8.2cm]{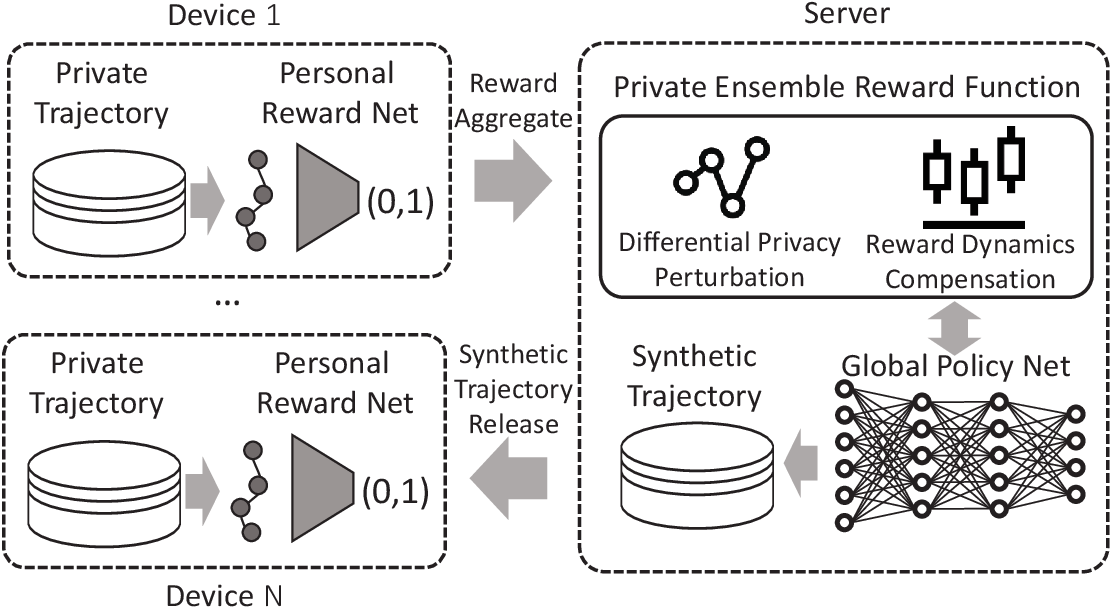}
\end{center}
\caption{The framework of our system.} \label{fig:framework}
\end{figure}

\subsection{Mathematical Model}

For the sake of convenience, we summarize the notations used in this paper in Table~A1 of the Appendix. Specifically, we consider the scenario where there are multiple users with their own mobile devices. Each device has recorded the historical mobility trajectory of the corresponding user. We define the set of users as $\mathcal{U}$. Further, for each user $u\in\mathcal{U}$, we define the mobility trajectory of $u$ as a sequence of spatio-temporal points, i.e., $T_u=\{(t_1,l_1),(t_2,l_2),...,(t_N,l_N)\}$, where $l_i$ is the identifier of the visited location and $t_i$ is the corresponding timestamp. Then, the human mobility generation problem can be defined as follows.

\begin{definition}[Privacy-Preserving Federated Mobility Trajectory Generation Problem]\label{def:prob}
Given a set of user $\mathcal{U}$ and their historical mobility trajectory $\{T_u\}_{u\in\mathcal{U}}$, the goal of this problem is to train a mobility trajectory generator distributedly, which is able to generate mobility trajectory with the preserved utility. In addition, the privacy information involved in the trajectory of each user should be preserved.
\end{definition}

Specifically, the preserved utility indicates that the generated trajectories should statistically resemble the real-world trajectories. Furthermore, they should be able to effectively support the downstream applications relying on trajectory data. On the other hand, the preserved privacy indicates that any pieces of the users' private trajectories should neither be taken away from their own devices, nor be inferred from the uploaded intermediate results of the user devices.

\subsection{System Overview}
We propose a privacy-preserving federated imitation learning system to solve this problem, of which the framework is shown in Figure~\ref{fig:framework}. 

As we can observe, each device keeps the private trajectory data belonging to its owner (user), and it trains a personal discriminator to measure to what degree arbitrary state-action pair $(s,a)$ resembles its owner. Thus, this discriminator is trained based on the positive samples obtained from the real-world user trajectory belonging to the user, while the negative samples come from the trajectory generator in our system. Then, only the reward obtained from users' personal discriminators is uploaded to the server.

Although the trajectories are not uploaded, user privacy is also possible to be leaked from the uploaded intermediate result through membership inference attacks~\cite{Shokri2017Membership} or reconstruction attack~\cite{geiping2020inverting}. Thus, a private aggregation mechanism is utilized to aggregate the reward obtained from users' personal discriminators for arbitrary state-action pair $(s,a)$. Specifically, a differential privacy perturbation mechanism is utilized in this process to protect users' privacy. Furthermore, a reward dynamics compensation based on the variance of the rewards is utilized to eliminate the potential noise introduced by the ensemble learning and model the dynamics of reward function across users. Finally, based on the obtained overall reward, the global policy network is able to be trained with the target of finding a policy to maximize the obtained rewards.

\section{Method}\label{sec:gail}

In order to model the substantial human decision-making process to simulate the human mobility behavior, we utilize the powerful technique of the generative adversary imitation learning (GAIL) under the model of Markov decision processes (MDPs).
Specifically, the MDP model is defined by a 4-tuple $<\mathcal{S},\mathcal{A},P,R>$, where $\mathcal{S}$ is the state space, $\mathcal{A}$ is the action space, $P:\mathcal{S}\times\mathcal{A}\times\mathcal{S}\rightarrow\mathbb{R}+$ represents the state transition probability, and $R:\mathcal{S}\times\mathcal{A}\rightarrow\mathbb{R}$ represents the reward function. 
Specifically, we define the state of users as the set of their historical spatio-temporal points, i.e., $s_t=\{(t_\tau,l_\tau)\}_{\tau\leq t}$, and the action space is defined based on the widely-adopted {\em exploration and preferential return (EPR)} model~\cite{jiang2016timegeo,song2010limits}, which includes four actions composed of {\em stay, home return, preferential return, and explore}. Then, the state transition probability, which defines the probability distribution of the next state given the current state and action, is also defined based on the EPR model (see Appendix for details).
Then, each user is regarded as an ``agent'' who dynamically determines the action to be executed based on its current state through its policy function, and the goal of the agent is to maximize the discounted total rewards $\sum_{t}\gamma^{t}R(s_t,a_t)$, where $\gamma\leq 1$ is the discount factor. 
In the imitation learning problem, the reward function $R$ as well as the policy function are unknown and thus need to be learned from the real-world data.
Thus, in the following part of this section, we first introduce the utilized policy function and reward function. Then, we introduce how to train our proposed system based on GAIL. Finally, we analyze the system in terms of its theoretical privacy-preserving performance.

\subsection{Policy Function}\label{sec:policy}

The policy function defines the decision strategy taken by users, which takes the current state $s_t$ as the input and then outputs the action $a_t$ to be executed. We follow the common settings adopted in most imitation learning problems, and consider the stochastic policy rather than the deterministic policy. In this case, the policy function actually gives the probabilistic distribution of executing arbitrary action $a_t\in\mathcal{A}$. Specifically, we utilize a self-attention transformer~\cite{vaswani2017attention} parameterized by $\theta$ to model the policy function, which is denoted by $\pi_\theta(a_t|s_t)$.

Combining the policy function $\pi(a_t|s_t)$ and the state transition probability function $P(s_{t+1}|s_t,a_t)$, each agent is able to dynamically determine which action to be executed and then change the current state from $s_t$ to $s_{t+1}$ via interaction with $P(s_{t+1}|s_t,a_t)$. By repeating this process, the agent is able to sample synthetic trajectories corresponding to the policy net $\pi_\theta$. Thus, the policy function acts as the trajectory generator in our system. 

In our system, we only utilize a global policy network, which is trained on the server.  Further, in the training process of our system, the policy network only interacts with the private user trajectories through the reward function. By designing a privacy-preserving reward function, we are able to obtain a policy network without privacy leakage. Then, the server is able to send the parameter of the global policy network to the devices, and thus the devices also have the ability to generate synthetic human trajectories. 
Specifically, the policy function is optimized to imitate real-world human trajectories, of which the degree is measured by the reward function introduced in the following section.

\subsection{Reward Function}\label{sec:reward}

The reward function measures to what degree arbitrary given trajectories imitate real trajectories. Specifically, it takes a state-action pair $(s_t,a_t)$ as the input, and outputs a real number, where higher values indicate that the state-action pair better imitates real-world human decisions.

In the standard GAIL, a discriminator network is utilized to model the reward function, which is trained based on the positive samples of real-world human state-action pairs and the negative samples of synthetic state-action pairs.
However, in order to protect user privacy, the private trajectory data stored on mobile devices cannot be gathered together to train the discriminator. Thus, in our system, we replace it with a number of separate personal discriminators of users and a private aggregation mechanism, of which the idea is inspired by the techniques of Private Aggregation of Teacher Ensembles (PATE)~\cite{papernot2016semi,jordon2018pate}.
Specifically, each user device trains its personal discriminator based on its private trajectory data, and the final utilized reward function comes from the private aggregation of their ensemble.
Note that the personal discriminators play a similar role with the teacher models in the standard PATE and PATE-GAN model~\cite{papernot2016semi,jordon2018pate}.
However, different from them, there is no student discriminator trained in our designed system. 
Instead, the aggregated reward is utilized to directly update the quality function or the advantage function in reinforcement learning algorithms such as A2C~\cite{mnih2016asynchronous} and PPO~\cite{schulman2017proximal}. 

In the following part of this section, we first introduce the personal discriminator. Then, we introduce how to implement a private aggregation to obtain the reward function based on ensemble learning. Finally, we propose a reward dynamics compensation mechanism to eliminate the potential noise introduced by ensemble learning and model the dynamics of reward function across users.

\para{Personal Discriminators:} The discriminator takes the state-action pair as input and then outputs its plausibility. The personal discriminator plays a similar role, but it can only access the trajectory data belonging to its corresponding user and the synthetic trajectories generated based on the global policy network. Specifically, we denote the personal discriminator belonging to user $u$ as $D_{\phi_u}$, which is parameterized by $\phi_u$. Then, $D_{\phi_u}$ is optimized based on the following loss function:
\begin{equation}\label{equ:teacherdisc}
   \mathcal{L}^u_D(\phi_u)=-\mathbb{E}_{\pi_{T_u}}[{\rm log}{D_{\phi_u}}(s,a)]-\mathbb{E}_{\pi}[{\rm log}(1-{D_{\phi_u}}(s,a))],
\end{equation}
where $\mathbb{E}_{\pi}$ represents the expectation with respect to the trajectories sampled based on the policy $\pi$.
Specifically, for an arbitrary function $f: \mathcal{S}\times\mathcal{A}\rightarrow \mathbb{R}$, we have $\mathbb{E}_{\pi}[f(s,a)]=\mathbb{E}[\sum^{N}_{i=1} f(s_i,a_i)]$, where $a_i\sim\pi(\cdot|s_i)$ and $s_{i+1}\sim P(\cdot|s_i,a_i)$.
In addition, $\mathbb{E}_{\pi_{T_u}}$ represents the expectation in terms of the state-action pairs obtained from the real-world trajectory of user $u$.

\para{Private Aggregation Mechanism:}
The trajectory data on each user device is insufficient and largely influenced by the user personality, which prevents the discriminator from capturing the principle plausibility of state-action pairs. Thus, it is necessary to incorporate the plausibility estimated by all personal discriminators. 

Formally, for arbitrary state-action pair $(s,a)$, each mobile device $u$ estimates its plausibility based on the local personal discriminator $D_{\phi_u}(s,a)$, which is then uploaded to the server.
The server computes the average value of the obtained personal rewards and then adds a perturbation to it, of which the process can be expressed as follows:
\begin{equation}\label{equ:pate}
{ R}(s,a)=\frac{1}{|\mathcal{U}|}\sum_{u\in\mathcal{U}}D_{\phi_u}(s,a)+{\rm Laplace }(0,\lambda).
\end{equation}
Note that this process from uploading  $ D_{\phi_u}(s,a)$ to calculating (\ref{equ:pate}) can be protected by homomorphic encryption techniques. Specifically, the server is in charge of generating public keys and sending them to all devices. Each device $u$ then encrypts $D_{\phi_u}(s,a)$ and sends it to another third-party server, which is in charge of implementing the computation of (\ref{equ:pate}) and sending the results to the server in charge of training the trajectory generator. 

The third-party server can only influence the performance of the trained trajectory generator and no user privacy can be leaked from it. To order to further prevent malicious third-party servers, we can randomly select a client to be the third-party server in each communication round.
It is also possible for the malicious third-party server to incorrectly calculate (\ref{equ:pate}), which can be solved by introducing verification information to the uploaded message $D_{\phi_u}(s,a)$ of the clients. 

\begin{table*}[t]
\centering

\label{table:performance}
\resizebox{1.0\textwidth}{!}{
\begin{tabular}{|c|c|c|c|c|c|c|c|c|c|c|}
\hline
\textbf{Dataset} & \multicolumn{5}{c|}{\textbf{ISP}} & \multicolumn{5}{c|}{\textbf{GeoLife}} \\ \hline
\textbf{Metrics(JSD)}& \emph{Radius} & \emph{DailyLoc} & \emph{Distance} & \emph{G-rank} & \emph{I-rank} & \emph{Radius} & \emph{DailyLoc} & \emph{Distance}& \emph{G-rank} & \emph{I-rank}  \\ \hline
\textbf{IO-HMM } & \ul{0.1443} & 0.3929 & 0.0596 & 0.0635 & 0.1005
 & 0.6146 & 0.6928 & 0.5108 & 0.1679 & 0.0529   \\ \hline
\textbf{TimeGeo} & 0.1609 & 0.6912 & 0.0337 & 0.0875 & 0.1125
& \ul{0.0737} & 0.5349 & 0.0473 & 0.0553 & 0.0584   \\ \hline
\textbf{DeepMove} & 0.6425 & 0.6934 & 0.4483 & 0.1947 & 0.2310
& 0.6754 & 0.4914 & 0.0512 & 0.1302 & 0.0934
  \\ \hline
\textbf{GAN}  & 0.6267 & 0.6936 & 0.4421 & 0.1022 & 0.2485
& 0.6143 & 0.6932 & 0.5157 & 0.0550 & 0.3000 \\ \hline
\textbf{SeqGAN} & 0.6297  & 0.6931 & 0.4388 & 0.1537 & 0.2474
& 0.6146 & 0.6927 & 0.5068 & 0.0535 & 0.2867 \\ \hline
\textbf{MoveSim}  & 0.3606 & \ul{0.1130} & \ul{0.0245} & \ul{0.0578} & \ul{0.0816}  
& 0.2845  & \ul{0.2467}&  \ul{0.0138} & \ul{0.0492} & \ul{0.0408}   \\ \hline
\textbf{Ours} & \textbf{0.0556}  &  \textbf{0.0381} & \textbf{0.0051} & \textbf{0.0510} & \textbf{0.0096} 
& \textbf{0.0699} & \textbf{0.1046} & \textbf{0.0130} & \textbf{0.0256} & \textbf{0.0176}  \\ \hline
\textbf{Percentages} & \textbf{61.47\%} & \textbf{66.28\%} & \textbf{79.18\%} & \textbf{11.76\%} &  \textbf{88.24\%}
& \textbf{5.16\%} & \textbf{57.60\%} & \textbf{5.80\%} & \textbf{47.97\%} & \textbf{56.86\%}    \\ \hline
\end{tabular}}
\caption{Performance comparison of our model and baselines on two mobility datasets, where lower results are better. Bold denotes best (lowest) results and underline denotes the second-best results.}
\end{table*}

Another difference of the aggregation mechanism (\ref{equ:pate}) from standard PATE is that we compute the average value of the uploaded outputs of the personal discriminators, while the aggregation results of the standard PATE are discrete. The reason is that utilizing a discrete aggregation mechanism may lead to a sparse reward function, which reduces the performance of our proposed system. Due to this difference, the required perturbation scale $\lambda$ to achieve differential privacy is also different.

\para{Reward Dynamics Compensation Mechanism:} The above ensemble learning based method will introduce extra noise to the obtained reward function. Specifically, users' mobility behavior has intrinsic stochasticity. Furthermore, there also exist personal differences between the reward functions of different users. Since each personal discriminator can only observe the trajectory of a single user, it is more affected by stochasticity and personal differences, which might reduce the performance.

In order to eliminate the influence introduced by the distributed training method, we propose a reward dynamics compensation mechanism. Specifically, it models the reward dynamics actively based on the variance of the obtained rewards from the personal discriminators, which is further incorporated into the reward function to derive the lower bound of obtained reward. The process can be formally expressed by the following equations:
\begin{equation}\label{equ:rlowerbound}
\begin{cases}
\xi(s,a)=\sqrt{{\rm var}(D_{\phi_u}(s,a))+{\rm Laplace }(0,\lambda_c)},\\
{\hat{R}}(s,a)={R}(s,a)-\beta\xi(s,a),\\
\end{cases}
\end{equation}
where ${\rm var}(X)$ is the variance of the stochastic variable $X$, and $\beta$ is a hyper-parameter to adjust the influence of the reward dynamics compensation mechanism. 
Intuitively, $\hat{R}$ can be regarded as the lower bound of the personal rewards of users, which is formally described in the following theorem.

\begin{mtheorem}\label{thm:coverge}
Denote the discounted total reward based on the policy function $\pi$ and reward function $R$ as $\boldsymbol{J}(\pi,R)=\sum_{i}\gamma^{i}R(s_i,a_i)$, where $a_i\sim\pi(\cdot|s_i)$ and $s_{i+1}\sim P(\cdot|s_i,a_i)$.
Let $R_u$ denote the personal reward function of user $u$, i.e., $R_u=D_{\phi_u}$.
Then, for a randomly selected user $u$ and policy function $\pi$,
we have $Pr\left(\boldsymbol{J}(\pi,R_u)\geq \boldsymbol{J}(\pi,\hat{R})\right)\geq 1-1/{\beta^2}$. 
\end{mtheorem}
{\em Proof.} See Appendix for proof. $\hfill\square$

This theorem tells us that under arbitrary policy function $\pi$, the discounted total reward of user $u$ obtained based on the reward function $\hat{R}$ is the lower bound of that obtained based on $R$ with probability $1-\frac{1}{\beta^2}$. By setting a sufficiently large $\beta$, we can obtain a probability very close to one.
Thus, by utilizing the reward function (\ref{equ:rlowerbound}) to replace the original reward function (\ref{equ:pate}), 
our proposed model is able to maximize the lower bound discounted total reward of the majority of users, which helps to eliminate the potential noise introduced by the ensemble learning and model the dynamics of reward function across users.

\subsection{Model Training} \label{sec:training}

The training process of our proposed model is summarized as follows. Firstly, a batch of synthetic trajectories is sampled to train the personal discriminators belonging to different users. Note that this generation process can be implemented on the server as well as the user devices by sending the parameters of the global generator to each device. Then, each user device samples a batch of positive samples from its private real-world trajectory data. Combined with the negative samples obtained from the synthetic trajectories, each device is able to optimize its own discriminator for a number of iterations.
Next, another batch of synthetic trajectories sampled from the server is sent to all devices. Each device calculates its rewards based on its personal discriminator and uploads the results to the server. After receiving the rewards of synthetic trajectories obtained from different devices, the server aggregates them based on (\ref{equ:pate})$~$(\ref{equ:rlowerbound}), which is used as the reward function of the imitation learning. Finally, based on the obtained overall reward, the server is able to optimize the global policy function based on reinforcement learning algorithms, e.g., PPO. We give the pseudocode of the training process in the Appendix.

\subsection{Privacy Analysis}\label{sec:privacy}
Before we analyze our proposed system in terms of its theoretical privacy-preserving performance, we first provide preliminaries about differential privacy, which is the most widely-used privacy preserving criterion.

\begin{definition}[$(\epsilon,\delta)$-differential privacy]\label{def:dp}
A randomized mechanism $\mathcal{M}: \mathcal{D} \rightarrow \mathcal{O}$ satisfies $(\epsilon,\delta)$-differential privacy if and only if, for arbitrary adjacent datasets $D_1$ and $D_2$, and subset $O\in\mathcal{O}$, we have $Pr(\mathcal{M}(D_1)\in O)\leq e^\epsilon Pr(\mathcal{M}(D_2)\in O)+\delta$.
\end{definition}

Then, based on the above definition, we will further examine the theoretical privacy-preserving performance of our system by investigating how to set the value of the parameters of the adding noise in (\ref{equ:pate}) and (\ref{equ:rlowerbound}) to achieve $(\epsilon,\delta)$-differential privacy. Overall, our obtained results can be summarized in the following theorems.

\begin{mtheorem}\label{thm:p1}
By setting $\lambda=\frac{1}{\epsilon|\mathcal{U}|}$, the random mechanism (\ref{equ:pate}) satisfy $(\epsilon,0)$-differential private.
\end{mtheorem}
{\em Proof.} See Appendix for proof. $\hfill\square$

Theorem~\ref{thm:p1} provides how to achieve $(\epsilon,\delta)$-differential privacy by using the reward function based on the private aggregation mechanism. If further utilizing the reward dynamics compensation mechanism, we can achieve $(\epsilon,\delta)$-differential privacy based on the following theorem.

\begin{mtheorem}\label{thm:p2}
By setting $\lambda=\frac{\kappa}{\epsilon|\mathcal{U}|}$ and $\lambda_c=\frac{3\kappa}{\epsilon(\kappa-1)|\mathcal{U}|}$ for all $\kappa> 1$, the random mechanism (\ref{equ:rlowerbound}) satisfy $(\epsilon,0)$-differential private.
\end{mtheorem}
{\em Proof.} See Appendix for proof. $\hfill\square$

From these theorems, we can observe that the minimum scales of the added perturbation to achieve $(\epsilon,0)$-differential privacy are all inversely proportional to the number of users participating in the learning process. Thus, we can only add a small perturbation with sufficient users, indicating our proposed system is feasible in practice.


\section{Experiments}

\subsection{Datasets}

We utilize two trajectory datasets to evaluate the performance of our proposed algorithm, which includes a publicly available dataset from previous work and a large-scale dataset obtained from an Internet service provider (ISP).

\para{ISP Dataset.} This dataset is provided by an Internet service provider (ISP), which records over 100,000 mobile users' access logs to different cellular base stations covering the duration of one week. Users' locations are obtained based on their accessed cellular base stations, while the timestamps of the access logs are also recorded, together composing the spatio-temporal mobility trajectories of the users.

\para{GeoLife Dataset.} This dataset is collected by \cite{zheng2010geolife}, which contains the mobility trajectories of 178 users, of which the duration is from April 2007 to October 2011. Users' locations are obtained from the GPS logs of their mobile phones, with each record containing the latitude, longitude, and timestamp.

\subsection{Experimental Settings}

\para{Compared Algorithms.}
In order to have a reliable evaluation of our proposed algorithm, we select the following state-of-the-art trajectory generation algorithms to be compared with:
(1) \textbf{{IO-HMM}}~\cite{yin2017generative} modifies the hidden Markov model to incorporate external context information, where the home and work locations of users are used as context information to generate synthetic trajectories.
(2) \textbf{{TimeGeo}}~\cite{jiang2016timegeo} is a rule-based probabilistic model based on the classical exploration and preferential return (EPR) model.
(4) \textbf{{GAN}}~\cite{goodfellow2014generative} utilizes the GAN model to directly generate trajectories, which trains the generator and the discriminator in an adversarial manner.
(5) \textbf{{SeqGAN}}~\cite{yu2017seqgan} modifies the standard GAN by enabling the generator to synthesize trajectories step by step, which is trained based on the policy gradient algorithm.
(6) \textbf{{MoveSim}}~\cite{FengYXYWL20} is a state-of-the-art trajectory generation algorithm based on GAN. Specifically, the domain knowledge of human mobility regularity is utilized to improve performance.

In addition, all baselines are implemented in the centralized setting, while only our proposed method is implemented in the distributed setting,
of which the detailed parameter settings are given in the Appendix.

\begin{figure}[t!]
\centering
\subfigure[]{\includegraphics[width=.234\textwidth]{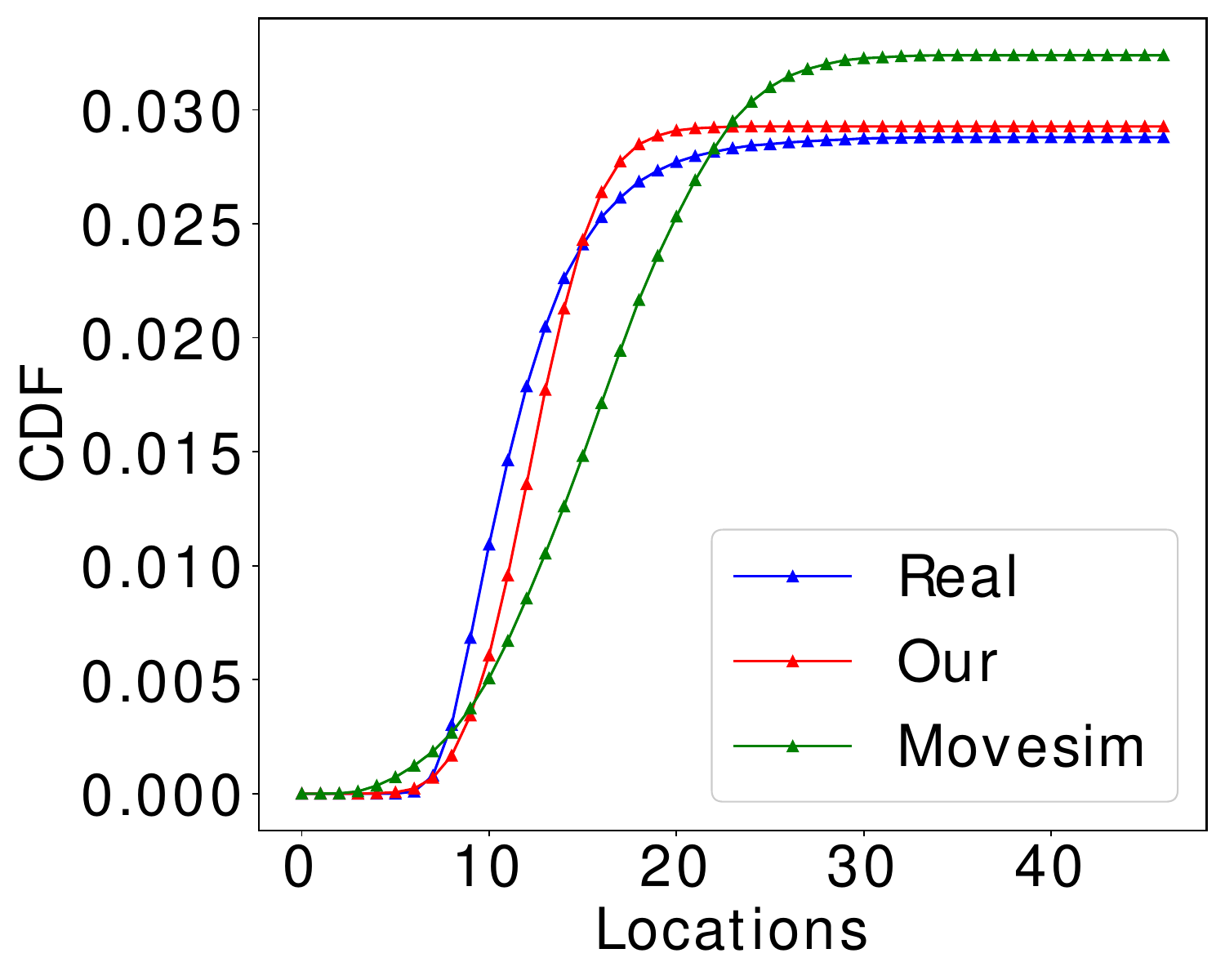}}
\subfigure[]{\includegraphics[width=.234\textwidth]{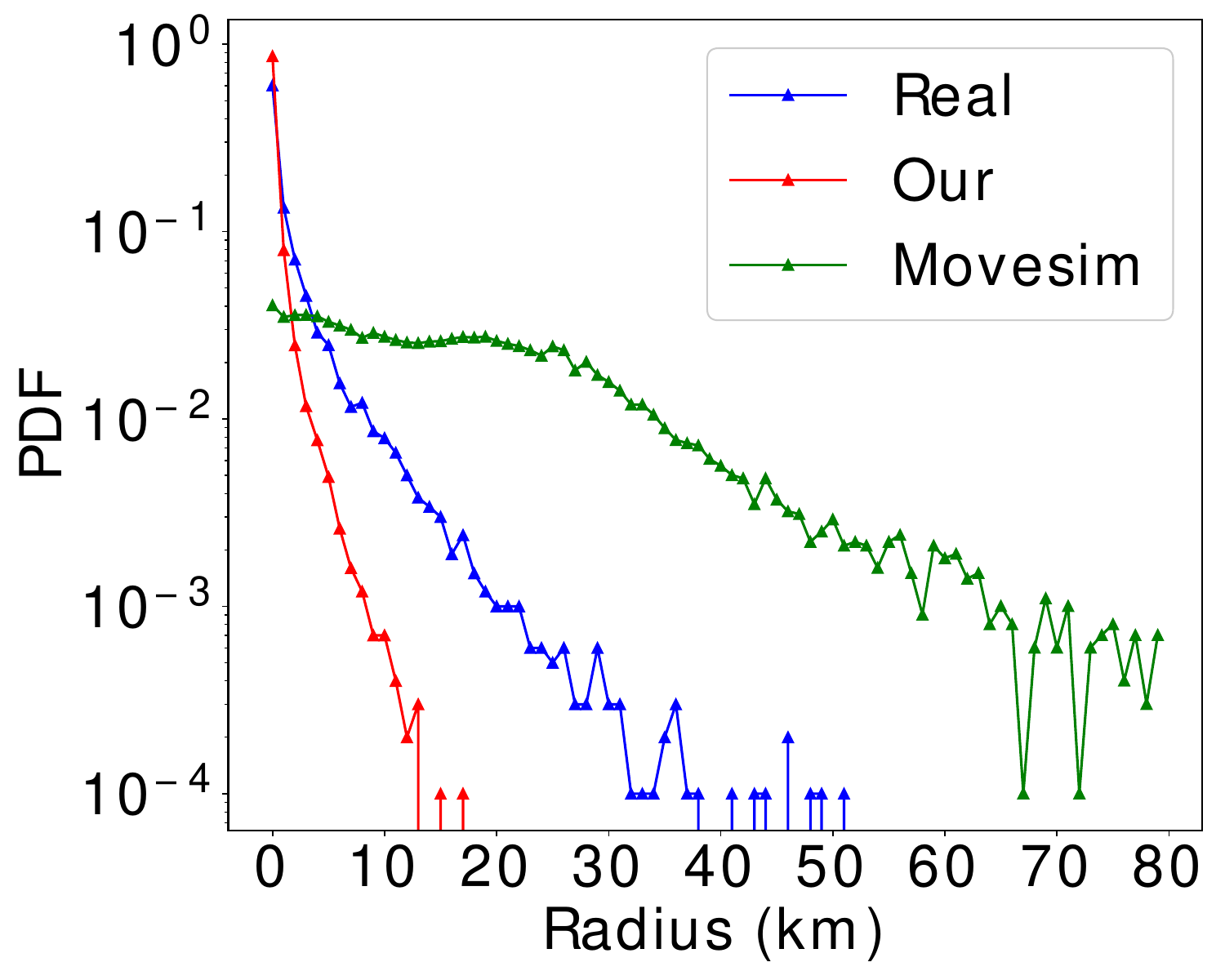}}\\
\subfigure[]{\includegraphics[width=.234\textwidth]{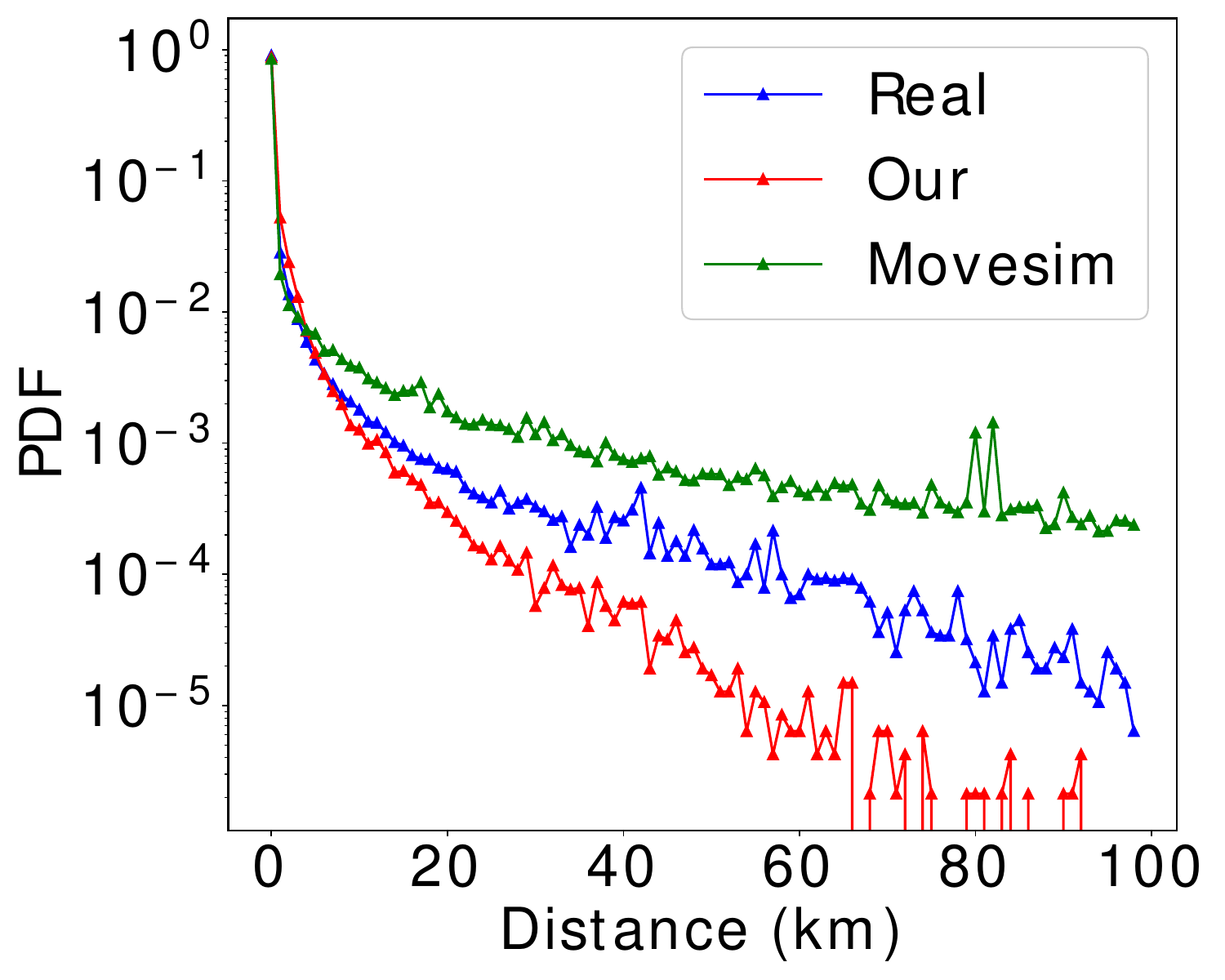}}
\subfigure[]{\includegraphics[width=.234\textwidth]{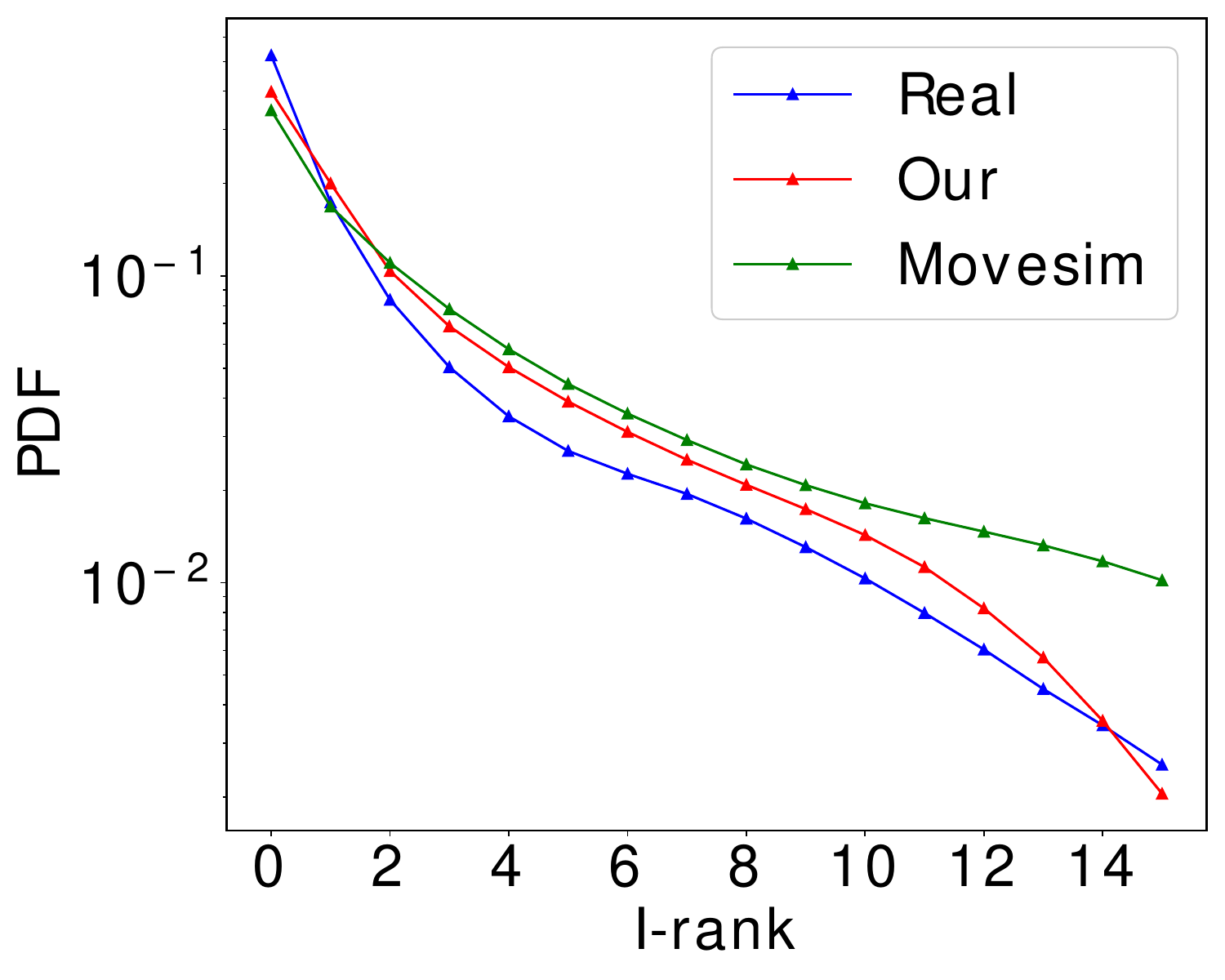}}
\caption{Visualization of the distribution of the selected statistical metrics on the ISP dataset.} \label{fig:dis_mobile}
\end{figure}

\para{Statistical Evaluation Metrics.}
In order to comprehensively evaluate the quality of the generated mobility trajectories, we first evaluate the similarity of the generated trajectories in terms of statistical characteristics. Specifically, we select the following indicators to characterize the characteristics of the mobility trajectories at the record-level or trajectory-level:
(1) \textbf{Radius} represents {\em radius of gyration}~\cite{Gonzalez2008Understanding}, i.e., a trajectory-level metric defined by the root mean square of each point of an arbitrary trajectory to its center of mass. 
(2) \textbf{DailyLoc} is a trajectory-level metric defined by the number of distinctive locations visited by each trajectory.
(3) \textbf{Distance} measures the traveling distance between the adjacent records in users' trajectories, and is a metric at the record-level.
(4) \textbf{G-rank} is a record-level metric defined by the normalized visited frequency of all users to the top visited locations.
(5) \textbf{I-rank} is also a record-level metric. Different from G-rank, this metric is computed based on the normalized visited frequency of each user to its top visited locations and then takes the average between different users.

Specifically, these metrics are all expressed by probability distributions with each mobility record or each trajectory as an example. In order to compare the generated trajectories and real trajectories in a more intuitive way, we utilize the Jensen-Shannon divergence (JSD) to measure their difference. Specifically, for two distributions $\boldsymbol{p}$ and $\boldsymbol{q}$, the JSD between them can be defined as:
\begin{equation}\label{equ:JSD}
{\rm JSD}(\boldsymbol{p},\boldsymbol{q})=\frac{1}{2}{\rm KL}(\boldsymbol{p}||\frac{\boldsymbol{p}+\boldsymbol{q}}{2})+\frac{1}{2}{\rm KL}(\boldsymbol{q}||\frac{\boldsymbol{p}+\boldsymbol{q}}{2}),
\end{equation}
where ${\rm KL}(\cdot||\cdot)$ is the Kullback-Leibler divergence~\cite{thomas2006elements}.

\subsection{Statistical Evaluation Results}

\para{Statistical Metrics.} We evaluate the performance of different algorithms in terms of statistical metrics. For fairness, in this group of experiments, no perturbation is added to our proposed algorithm. As the results shown in Table~\ref{table:performance}, our model beats the baselines in most situations. Compared to the best baseline, our method can obtain a significant performance gap in most metrics. In addition, we can observe that the JSD of most metrics of all algorithms on the GeoLife data is larger than those on the ISP dataset, indicating worse performance. The reason is that the GeoLife dataset is sparser and with a smaller scale than the ISP dataset, leading to the difficulty of capturing the complex temporal and spatial features by only relying on the limited number of trajectories.

\para{Statistical Distribution Visualization.} In Figure~\ref{fig:dis_mobile}, we compare the distribution of trajectory data generated from our proposed model and MoveSim on the ISP dataset in terms of the selected statistics metrics. We can observe that the distributions of trajectory data generated by our model are closer to the distribution of real data compared with MoveSim. We also present the corresponding experimental results of the GeoLife dataset in the Appendix.

\subsection{Practical Demonstrations}

Due to privacy concerns and the collection cost, the available real trajectories are usually limited, which has become the bottleneck of performance of the machine learning models of the downstream applications. 
In this group of experiments, we examine whether the synthetic trajectory data can help to solve the problem of data scarcity in terms of realistic downstream applications, which include mobility prediction and location recommendation.

\begin{figure}[t!]
\centering
\subfigure[ISP]{\includegraphics[width=.234\textwidth]{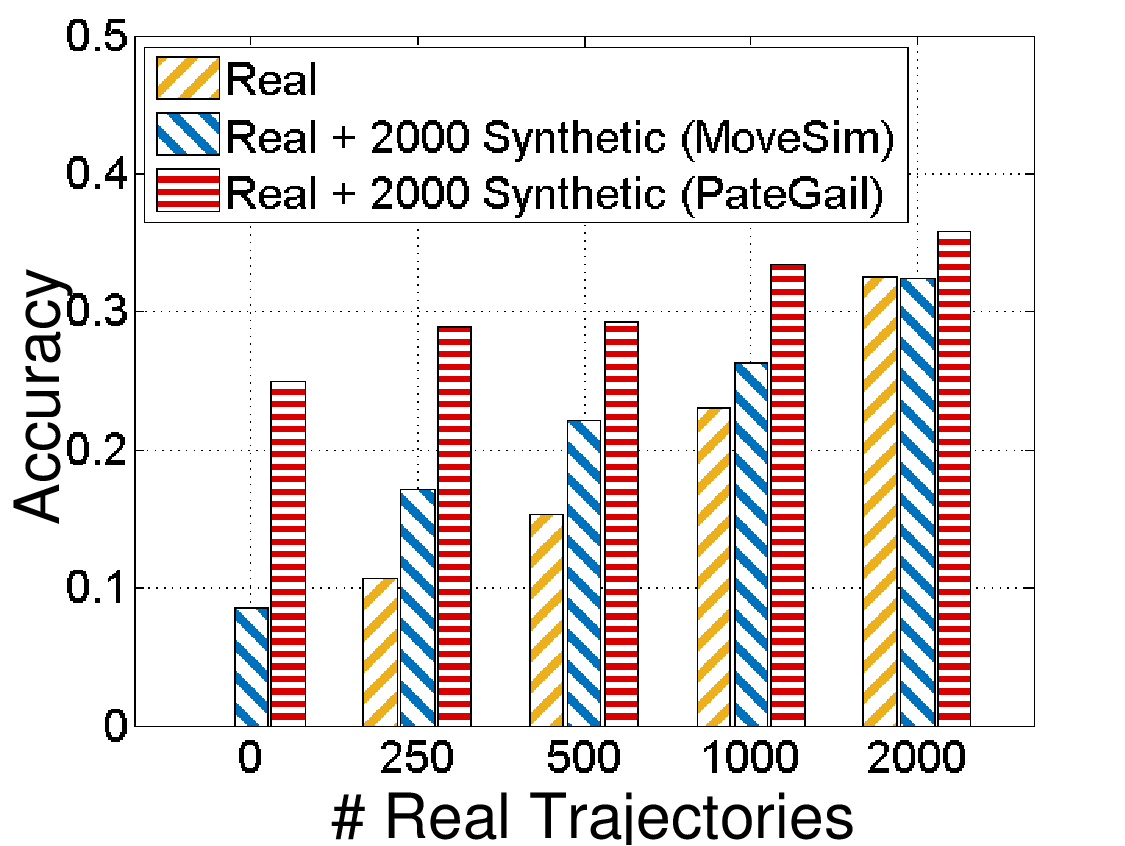}}
\subfigure[Geolife]{\includegraphics[width=.234\textwidth]{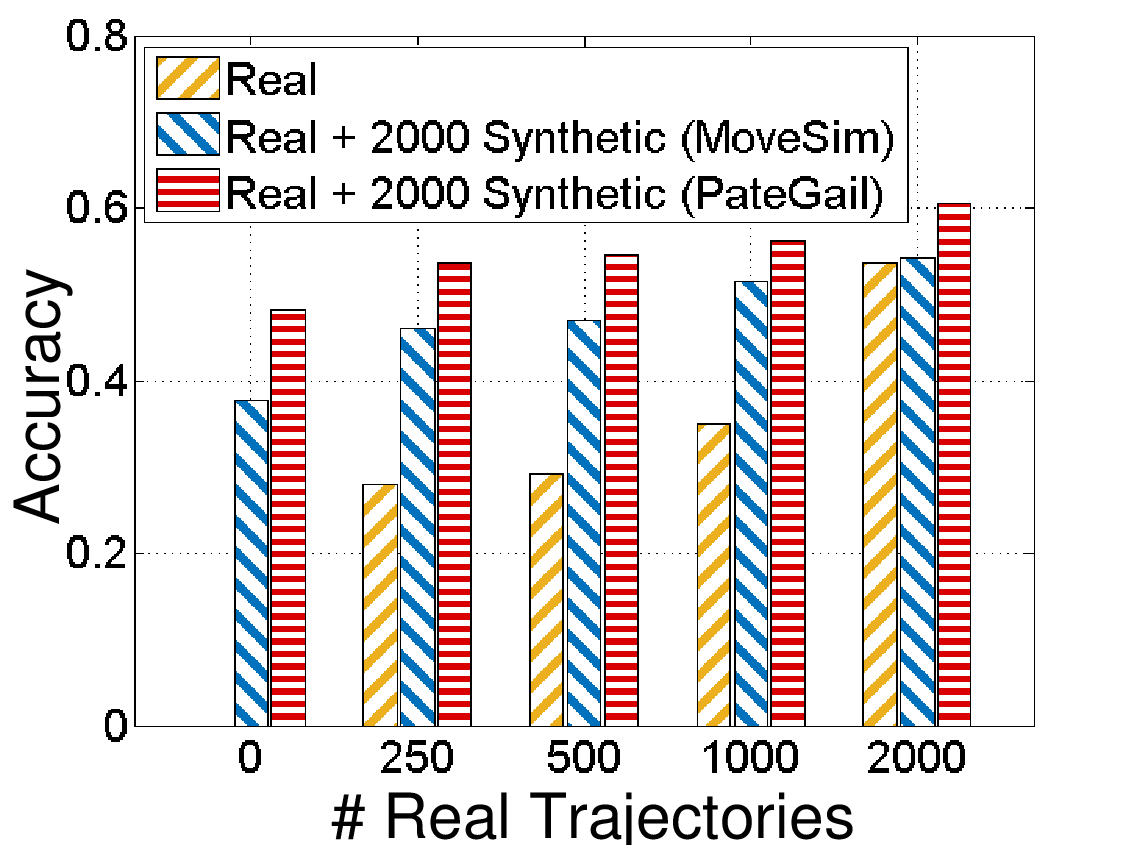}}
\caption{Performance of the individual mobility prediction based on trajectory
data augmented by different models.} \label{fig:movepred}
\end{figure}

\para{Mobility Prediction.}
In this experiment, we examine whether the synthetic trajectories can help to train a better mobility prediction model. Specifically, the real trajectories combined with synthetic trajectories are used to train an LSTM mobility prediction model~\cite{fattore2020automec}, which is then validated on another group of real trajectories as the test set.
We consider three different scenarios which only use real-world trajectory data, use real-world trajectory data and synthetic data generated by MoveSim, and use real-world trajectory data and synthetic data generated by our proposed algorithms as the training set, respectively.
We select MoveSim since it is the best baseline in the previous analysis.
As we can observe from Figure~\ref{fig:movepred}, the mobility prediction performance has a significant improvement when adding the generated trajectories, and the improvement of adding trajectories generated by our proposed algorithm is significantly greater than MoveSim. In addition, we also evaluate the performance when only using generated data, and our performance is approximately three times of MoveSim. These experiment results prove the usability of our proposed model.

\begin{figure}[t!]
\centering
\subfigure[ISP]{\includegraphics[width=.234\textwidth]{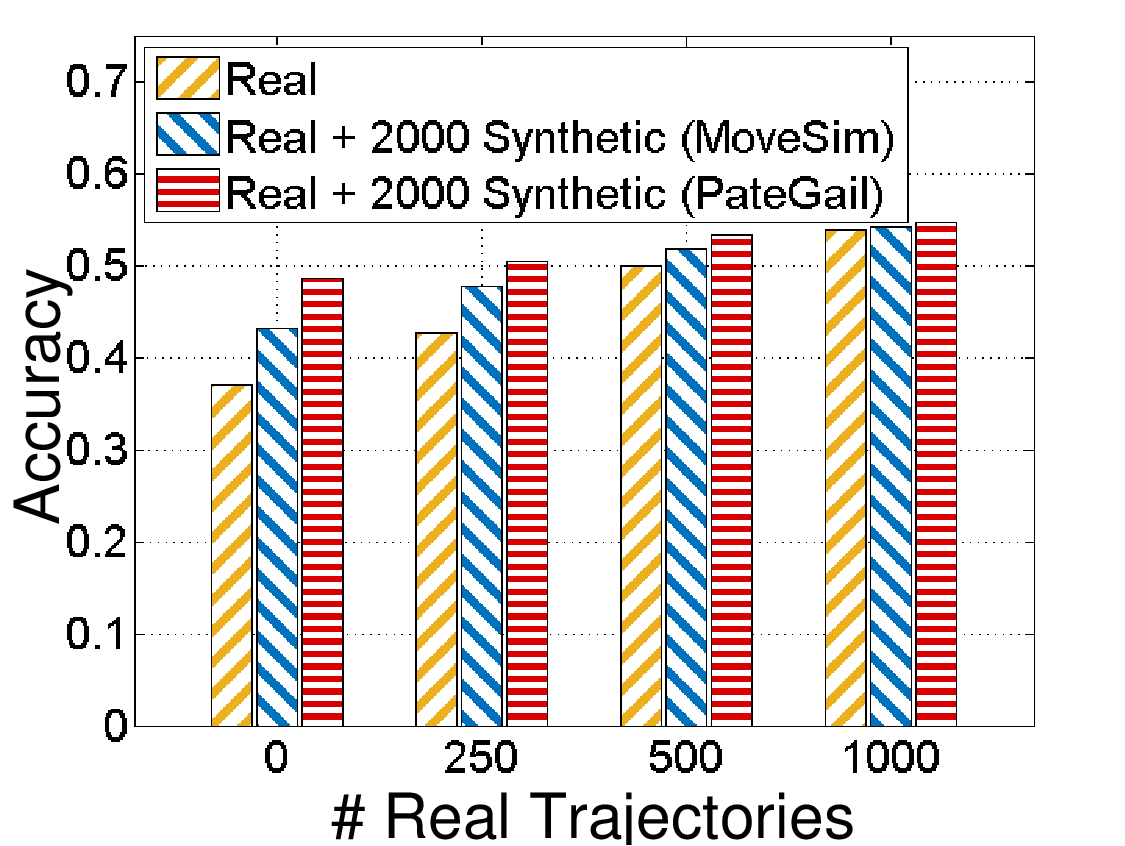}}
\subfigure[Geolife]{\includegraphics[width=.234\textwidth]{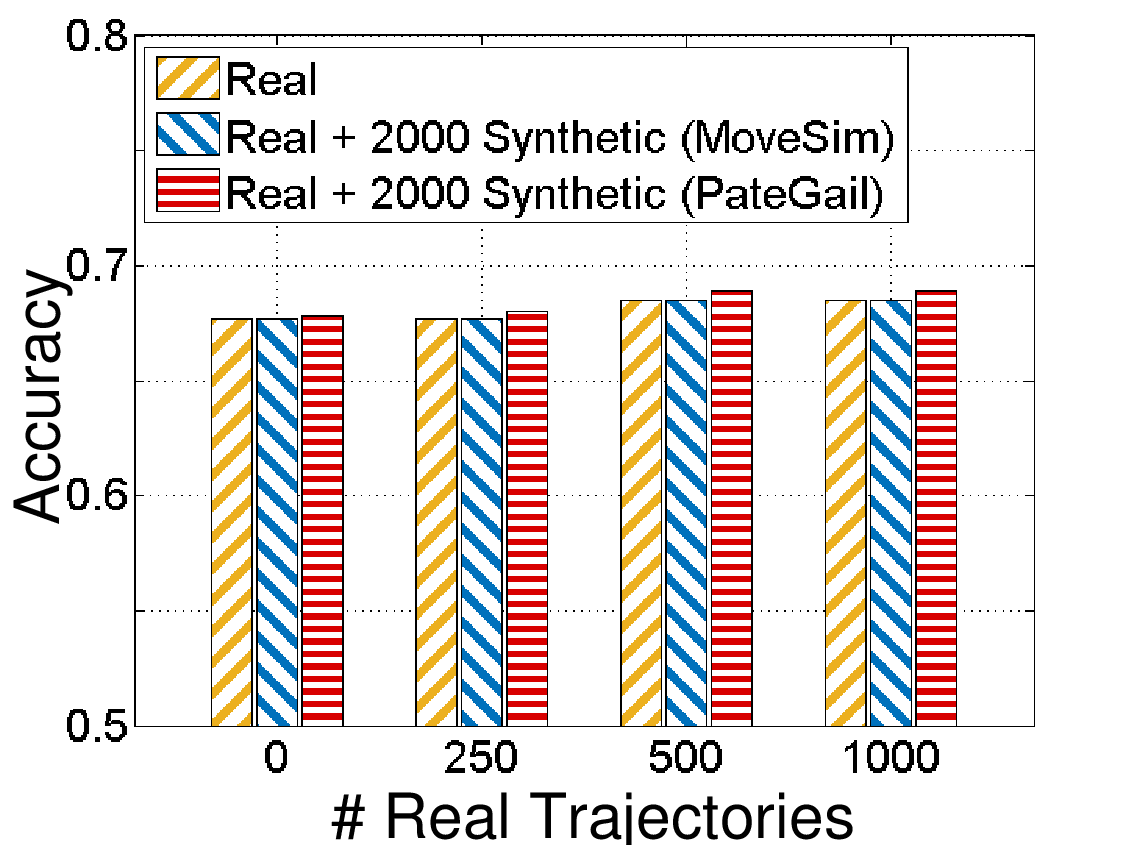}}
\caption{Location recommendation on different datasets.} \label{fig:locrec}
\end{figure}

\para{Location Recommendation.} 
Further, we utilize synthetic trajectories to train a collaborative filtering algorithm~\cite{he2016fast}, which is then utilized to recommend locations to real users.
As we can observe from Figure~\ref{fig:locrec}, although the performance gap is not significant on the GeoLife dataset due to its sparsity, on the ISP dataset, the performance is greatly improved by adding generated data. In addition, the large performance gap of the location recommendation algorithm based on trajectories generated by our proposed algorithm compared with MoveSim indicates the superiority of our proposed algorithm.

\subsection{Privacy Risk Analysis}

\begin{figure}[t!]
\centering
\subfigure[MIA]{\includegraphics[width=.234\textwidth]{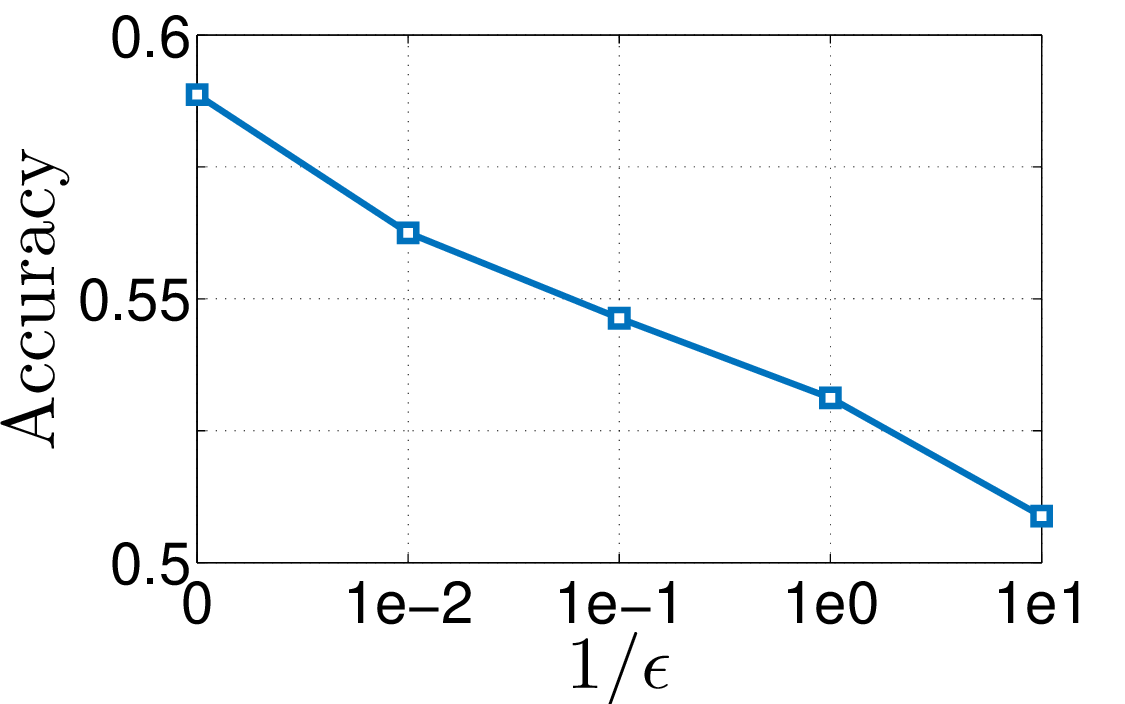}}
\subfigure[Uniqueness test]{\includegraphics[width=.234\textwidth]{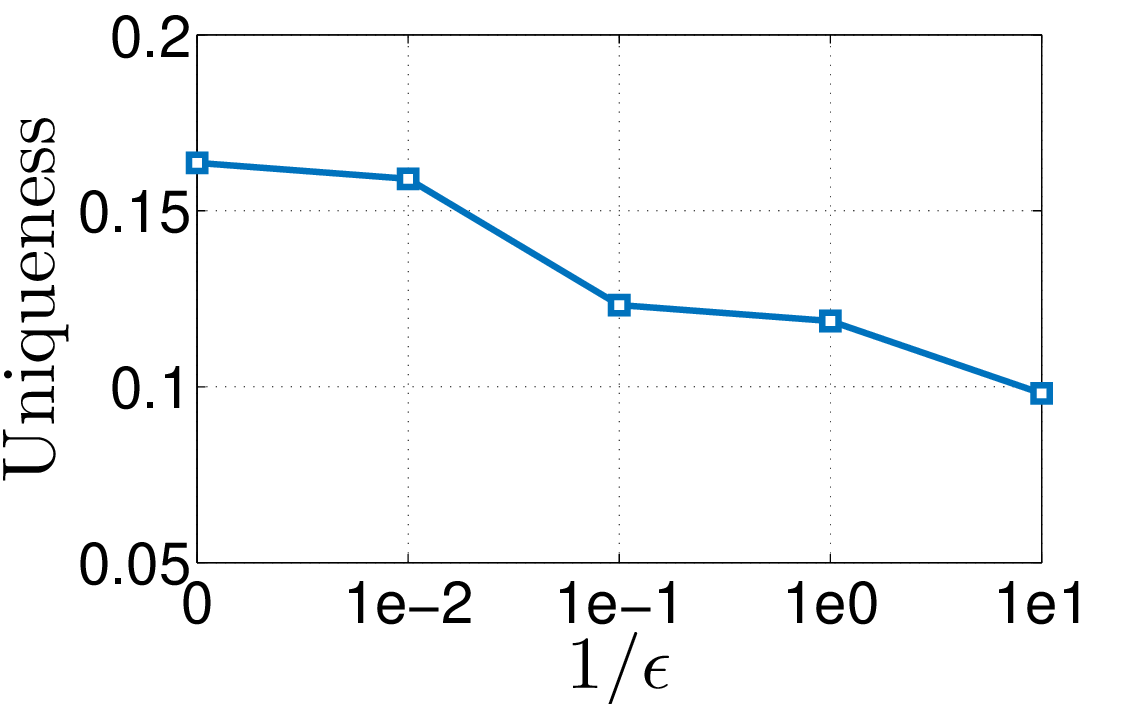}}
\caption{Performance in terms of privacy risk.} \label{fig:attack}
\end{figure}

To evaluate the privacy-preserving performance of our proposed algorithms, we further conduct two experiments in terms of actual privacy attacks.
The first attack is the membership inference attack (MIA)~\cite{Shokri2017Membership,nasr2019comprehensive}. Specifically, we consider the white-box inference attack. 
Given a set of trajectories $\mathcal{T}_A$ as the target of attacks, for each trajectory $T_u\in\mathcal{T}_A$, the adversary calculates the reward of each state-action pair in $T_u$, and then uses them as the feature of a Random Forest classifier to infer whether $T_u$ is included in the training dataset of the obtained trajectory generation model. 
The same number of real-world trajectories used for training and not used for training the trajectory generation model are sampled as the positive samples and  negative samples, respectively. Then, a five-fold cross validation is implemented to evaluate the privacy risk in terms of MIA.
The second attack is to examine the uniqueness of real trajectories with respect to generated trajectories. Specifically, for each real trajectory $T_u$, the adversary finds the generated trajectory $T'_v$ with the highest overlapping rate with $T'_v$, where the highest overlapping rate is utilized as the uniqueness metric.
Here, the overlapping rate between two trajectories is defined as the ratio between their identical locations at the same time slots and the total trajectory length.
A lower accuracy of MIA and a smaller uniqueness indicate better privacy-preserving performance. 
The experimental results are shown in Figure~\ref{fig:attack}.
As we can observe, it is not surprising that both MIA accuracy and uniqueness decrease with $1/\epsilon$, indicating less privacy risk for smaller privacy budget $\epsilon$.
In addition, we can observe that our algorithm has low MIA accuracy and uniqueness, indicating that it is robust to actual privacy attacks.

\section{Related Work}

\para{Human Trajectory Generators.}
Human trajectory generation models have been investigated for decades. Early approaches mainly utilize classical probabilistic methods or rule-based methods to model and generate human trajectories~\cite{jiang2016timegeo,isaacman2012human,yin2017generative,bindschaedler2016synthesizing,Zhao2019Synthesizing,Pappalardo2017Data}. 
However, these methods are derived from strong assumptions of human mobility, and whether these assumptions hold true in reality is questionable. In addition, these assumptions also limit a small number of parameters describing the human mobility process, leading to their weakness in terms of modeling the complicated relationship of high-dimensional mobility trajectories. In recent years, more deep learning based human trajectory generators have been proposed by utilizing variational autoencoder (VAE)~\cite{huang2019variational}, generative adversarial network (GAN)~\cite{feng2018deepmove,ouyang2018non,kulkarni2018generative,Liu2018trajGANsU,SocialGAN}, imitation learning~\cite{pan2020xgail,zhang2019unveiling,wu2020joint, ZangTrajGAIL, SeongjinChoi, ISrein}. However, none of them considers the critical privacy problem, indicating that there exist privacy leakage risks of the trajectory data utilized to train these models. Different from them, in this paper, we address the privacy-preserving issue, and propose a federated mobility generator by utilizing the techniques of imitation learning.

\para{Imitation Learning.} 
The goal of imitation learning is to learn the policy function, which gives the action to be executed based on the current state~\cite{bain1995framework,boularias2011relative,ziebart2008maximum,ziebart2010modeling,ziebart2008maximum,ho2016generative}. 
The most successful imitation learning method is generative adversarial imitation learning (GAIL), which utilizes the non-linear neural network to model the reward function and policy function. It has been adopted in numerous practical applications, including dynamic treatment regimes~\cite{Wang2020Adversarial}, traffic signal control~\cite{Xiong2019learning}, and human drive behavior analysis~\cite{pan2020xgail,zhang2019unveiling,wu2020joint, ZangTrajGAIL, SeongjinChoi, ISrein}, etc.  In this paper, we utilize imitation learning techniques to solve the human mobility trajectory generation problem, which is able to model the crucial human decision-making process to generate human trajectories with preserved utility.

\section{Conclusion}
In this paper, we propose a privacy-preserving federated mobility trajectory generator based on imitation learning techniques, which is able to generate plausible synthetic mobility trajectories with the preserved utility to be utilized in downstream applications and preserve users’ privacy at the same time.
Extensive experiments validate the effectiveness of our proposed model. Specifically, the generated trajectories based on our proposed algorithm are able to preserve the statistical properties of the original dataset in terms of a number of key statistical metrics.
Furthermore, the synthetic trajectories are able to efficiently support practical applications, including mobility prediction and location recommendation, demonstrating its effectiveness.

\bibliographystyle{plain}
\bibliography{bibliography}

\clearpage
\appendix
\setcounter{table}{0}   
\setcounter{figure}{0}
\renewcommand{\thetable}{A\arabic{table}}
\renewcommand{\thefigure}{A\arabic{figure}}

\section{Proofs of the Theorems}

\setcounter{mtheorem}{0}

\begin{mtheorem}
Denote the discounted total reward based on the policy function $\pi$ and reward function $R$ as $\boldsymbol{J}(\pi,R)=\sum_{i}\gamma^{i}R(s_i,a_i)$, where $a_i\sim\pi(\cdot|s_i)$ and $s_{i+1}\sim P(\cdot|s_i,a_i)$.
Let $R_u$ denote the personal reward function of user $u$, i.e., $R_u=D_{\phi_u}$.
Then, for a randomly selected user $u$ and policy function $\pi$,
we have $Pr\left(\boldsymbol{J}(\pi,R_u)\geq \boldsymbol{J}(\pi,\hat{R})\right)\geq 1-\frac{1}{\beta^2}$.
\end{mtheorem}

\textbf{Proof:}
First, based on Chebyshev's inequality, we have:
\begin{equation}\label{equ:cheby}
\begin{aligned}
 & Pr\left(\boldsymbol{J}(\pi,R_u)-\frac{1}{|\mathcal{U}|}\sum_{u\in\mathcal{U}}\boldsymbol{J}(\pi,R_u) \leq - \beta \sum_i \gamma^i\xi(s_i,a_i) \right) &  \\
 \leq & Pr\left(|\boldsymbol{J}(\pi,R_u)-\frac{1}{|\mathcal{U}|}\sum_{u\in\mathcal{U}}\boldsymbol{J}(\pi,R_u)| \geq  \beta \sum_i \gamma^i\xi(s_i,a_i) \right) &  \\
\leq  &   {\rm var}(\boldsymbol{J}(\pi,R_u))/
\left(\beta \sum_i \gamma^i\xi(s_i,a_i)\right)^2\\
\end{aligned}
\end{equation}

Then, in terms of $\boldsymbol{J}(\pi,R_u)$, since the policy function $\pi$ has been given, conditioned on $\pi$ the sampled station-action pair is independent with the reward function. Thus, we have $R_u(s_i,a_i)$ is independent with $R_u(s_j,a_j)$ for $i\not=j$. Then, we have:
\begin{equation}
\begin{aligned}
 & {\rm var}(J(\pi,R_u)) &  \\
 = & \frac{1}{|\mathcal{U}|}\sum_{u\in\mathcal{U}}\left[ \sum_i\gamma^i\left(R_u(s_i,a_i)- \bar{R}_u(s_i,a_i)\right)\right]^2  \\
=  &   \frac{1}{|\mathcal{U}|}\sum_{u\in\mathcal{U}}\left[ \sum_i\gamma^{2i}\left(R_u(s_i,a_i)- \bar{R}_u(s_i,a_i)\right)^2\right]\\
= & \sum_i\gamma^{2i}\xi^2(s_i,a_i)
\end{aligned}
\end{equation}

Since $\xi(s_i,a_i)\geq 0$ for all state-action pair $(s_i,a_i)$, we have $\sum_i\gamma^{2i}\xi^2(s_i,a_i)\leq \left(\sum_i \gamma^i\xi(s_i,a_i)\right)^2$. Then, based on (\ref{equ:cheby}), we have:
\begin{equation}
\footnotesize
  Pr\left(\boldsymbol{J}(\pi,R_u) \geq \frac{1}{|\mathcal{U}|}\sum_{u\in\mathcal{U}}\boldsymbol{J}(\pi,R_u) - \beta \sum_i \gamma^i\xi(s_i,a_i) \right) \geq 1-\frac{1}{\beta^2}.\ \
\end{equation}

At the same time, we have
\begin{equation}
   \boldsymbol{J}(\pi,\hat{R}) = \frac{1}{|\mathcal{U}|}\sum_{u\in\mathcal{U}}\boldsymbol{J}(\pi,R_u) - \beta \sum_i \gamma^i\xi(s_i,a_i),
\end{equation}
which proves our theorem. $\hfill\square$

\begin{mtheorem}
By setting $\lambda=\frac{1}{\epsilon|\mathcal{U}|}$, the query based on (\ref{equ:pate}) satisfy $(\epsilon,0)$-differential private.
\end{mtheorem}

\textbf{Proof:}
In our scenarios, the complete dataset is separately stored in $\mathcal{U}$ devices. Further, for an arbitrary query for the reward of the state-action pair $(s,a)$, we define the personal discriminator and the aggregated reward function (\ref{equ:pate}) conditioned on the complete dataset $D$ as $D_{\phi_u}(s,a|D)$ and $R(s,a|D)$, respectively.
Without loss of generality, for two arbitrary adjacent datasets $D_1$ and $D_2$, we assume they only differ in the trajectory data of the user $u_0$. Thus, we have
$
D_{\phi_u}(s,a|D_1)=D_{\phi_u}(s,a|D_2), \forall u\not=u_0
$.
Then, we have the following equations:
\begin{equation}
\setlength{\abovedisplayskip}{0pt}
\setlength{\belowdisplayskip}{0pt}
\footnotesize
\begin{aligned}
  \frac{p(R(s,a|D_1)=r)}{p(R(s,a|D_2)=r)}   = & \frac{{\rm exp}(-\frac{|\frac{1}{|\mathcal{U}|}\sum_{u\in\mathcal{U}}D_{\phi_u}(s,a|D_1)-r|}{\lambda})}{{\rm exp}(\frac{|\frac{1}{|\mathcal{U}|}\sum_{u\in\mathcal{U}}D_{\phi_u}(s,a|D_2)-r|}{\lambda})} \\
\leq  &  {\rm exp}(\frac{\frac{1}{|\mathcal{U}|}|\sum_{u\in\mathcal{U}}D_{\phi_u}(s,a|D_1)\!-\!\sum_{u\in\mathcal{U}}D_{\phi_u}(s,a|D_2)|}{\lambda})\\
=  &  {\rm exp}(\frac{\frac{1}{|\mathcal{U}|}|D_{\phi_{u_0}}(s,a|D_1)-D_{\phi_{u_0}}(s,a|D_2)|}{\lambda})\\
\leq  &  {\rm exp}(\frac{1/|\mathcal{U}|}{\lambda}) \\
=  & {\rm exp}(\epsilon).\\
\end{aligned}
\end{equation}

Then, for arbitrary set $O$, we have :
\begin{equation}
\setlength{\abovedisplayskip}{0pt}
\setlength{\belowdisplayskip}{0pt}
\begin{aligned}
  Pr(R(s,a|D_1)\in O)   = & \int_{r\in O}  p(R(s,a|D_1)=r){\rm d}r  \\
 \leq   &  \int_{r\in O}{\rm exp}(\epsilon)p(R(s,a|D_2)=r){\rm d}r\\
=  &  {\rm exp}(\epsilon)\int_{r\in O}p(R(s,a|D_2)=r){\rm d}r\\\\
\leq  &  {\rm exp}(\epsilon)Pr(R(s,a|D_2)\in O),
\end{aligned}
\end{equation}
which proves our theorem. $\hfill\square$

Before we prove Theorem~\ref{thm:p2}, we first provide some preliminaries in terms of lemmas that have been proven in existing studies.

\begin{mlemma}[Composition Theorem for Differential Privacy]\label{lem:composition}
For arbitrary randomized mechanism $\mathcal{M}_1$ satisfying $(\epsilon_1,\delta_1)$-differential private and  $\mathcal{M}_2$ satisfying $(\epsilon_2,\delta_2)$-DP, their combination $\mathcal{M}=(\mathcal{M}_1,\mathcal{M}_2)$ satisfying $(\epsilon_1+\epsilon_2,\delta_1+\delta_2)$-differential private.
\end{mlemma}

The proof of Lemma~\ref{lem:composition} can be found in \cite{dwork2014algorithmic}. Based on this lemma, we can prove Theorem~\ref{thm:p2} as follows.

\begin{mtheorem}
By setting $\lambda=\frac{\kappa}{\epsilon|\mathcal{U}|}$ and $\lambda_c=\frac{3\kappa}{\epsilon(\kappa-1)|\mathcal{U}|}$ for all $\kappa> 1$, the query based on (\ref{equ:rlowerbound}) satisfy $(\epsilon,0)$-differential private.
\end{mtheorem}
\textbf{Proof:}
Based on Theorem~\ref{thm:p1}, we can obtain that by setting $\lambda=\frac{\kappa}{\epsilon|\mathcal{U}|}$, the randomized mechanism $R(s,a)$ satisfies $(\epsilon/\kappa,0)$-differential private. Next, we further analyze the privacy bound of the randomized mechanism $\xi(s,a|D)$. Without loss of generality, we analyze the privacy bound of $\xi^2(s,a|D)$. Similarly, for two arbitrary adjacent datasets $D_1$ and $D_2$,  we have the following equations:
\begin{equation}
\setlength{\abovedisplayskip}{0pt}
\setlength{\belowdisplayskip}{0pt}
\scriptsize
\begin{aligned}
 & \frac{p(\xi^2(s,a|D_1)=o)}{p(\xi^2(s,a|D_2)=o)}   \\
 =   &  {\rm exp}(-\frac{|{\rm var}(D_{\phi_u}(s,a|D_1))-o|-|{\rm var}(D_{\phi_u}(s,a|D_2))-o|}{\lambda_c})\\
\leq  &  {\rm exp}(\frac{|{\rm var}(D_{\phi_u}(s,a|D_1))-{\rm var}(D_{\phi_u}(s,a|D_2))|}{\lambda_c})\\
\leq  &  {\rm exp}(\frac{      \frac{1}{|\mathcal{U}|}\left|D^2_{\phi_{u_0}}(s,a|D_1)-D^2_{\phi_{u_0}}(s,a|D_2)\right|     }{\lambda_c})\\
&\!\!\!\!\!\!\!\!\!\!\!\! \cdot {\rm exp}(\frac{      \left|D_{\phi_{u_0}}(s,a|D_1)-D_{\phi_{u_0}}(s,a|D_2)\right|  \cdot  \left|\overline{D_{\phi_{u_0}}(s,a|D_1)} + \overline{D_{\phi_{u_0}}(s,a|D_2)}\right|     }{\lambda_c|\mathcal{U}|})\\
\leq  &  {\rm exp}(\frac{3/|\mathcal{U}|}{\lambda_c})
=  {\rm exp}(\frac{3}{|\mathcal{U}|}\frac{\epsilon(\kappa-1)|\mathcal{U}|}{3\kappa})
=  {\rm exp}(\frac{\epsilon(\kappa-1)}{\kappa}).&
\end{aligned}
\end{equation}
Then, similarly with Theorem~\ref{thm:p1}, for arbitrarily $O\in\mathcal{O}$, we can have
\begin{equation}
\begin{aligned}
  Pr(\xi^2(s,a|D_1)\in O)
\leq  &  {\rm exp}(\frac{\epsilon(\kappa-1)}{\kappa})Pr(\xi^2(s,a|D_2)\in O). \\
\end{aligned}
\end{equation}
Thus, the random mechanism $\xi^2(s,a|D)$ satisfies $(\frac{\epsilon(\kappa-1)}{\kappa},0)$-differential private.
Further, $\hat{R}(s,a|D)$ can be regarded as the composition of the random mechanism $R(s,a|D)$ and $\xi^2(s,a|D)$. Thus, based on Theorem~\ref{lem:composition}, we can prove that $\hat{R}(s,a|D)$ satisfies $(\frac{\epsilon(\kappa-1)}{\kappa}+\frac{\epsilon}{\kappa},0)$-differential private, i.e, $(\epsilon,0)$-differential private. $\hfill\square$

\section{Details of Implementation}

\subsection{Notations and Pseudocode}

The notations used in this paper are summarized in Table~\ref{tab:symbal} and Algorithm~\ref{alg:framework} gives the pseudocode of the training process.

\subsection{Action Space and State Transition Probability}

The action space $\mathcal{A}$ in our system includes four actions, i.e., {\em stay, home return, preferential return}, and {\em explore}. Specifically, {\em stay} indicates that the user will not move to other locations in this time slot. {\em Home return} indicates that the user will go home in the next time slot. {\em Preferential return} indicates that the user will go to a previously visited location other than his/her home, while {\em explore} indicates that the user will go to a new location.

As for the state transition probability, the next states corresponding to the actions of {\em stay} and {\em home return} are deterministic. As for the next state corresponding to the action of {\em preferential return}, the newly visited location will be chosen based on the user's historical visitation frequency to this location. As for the next state corresponding to the action of {\em preferential return}, each location $l$ will be visited with the probability in proportion to ${\rm rank}(l)^{-\alpha}$, where ${\rm rank}(l)$ is the rank of the location $l$ based on its distance from the user's current location~\cite{jiang2016timegeo}.


\begin{figure}[t!]
\vspace{-0.2cm}
\centering
\subfigure[]{\includegraphics[width=.22\textwidth]{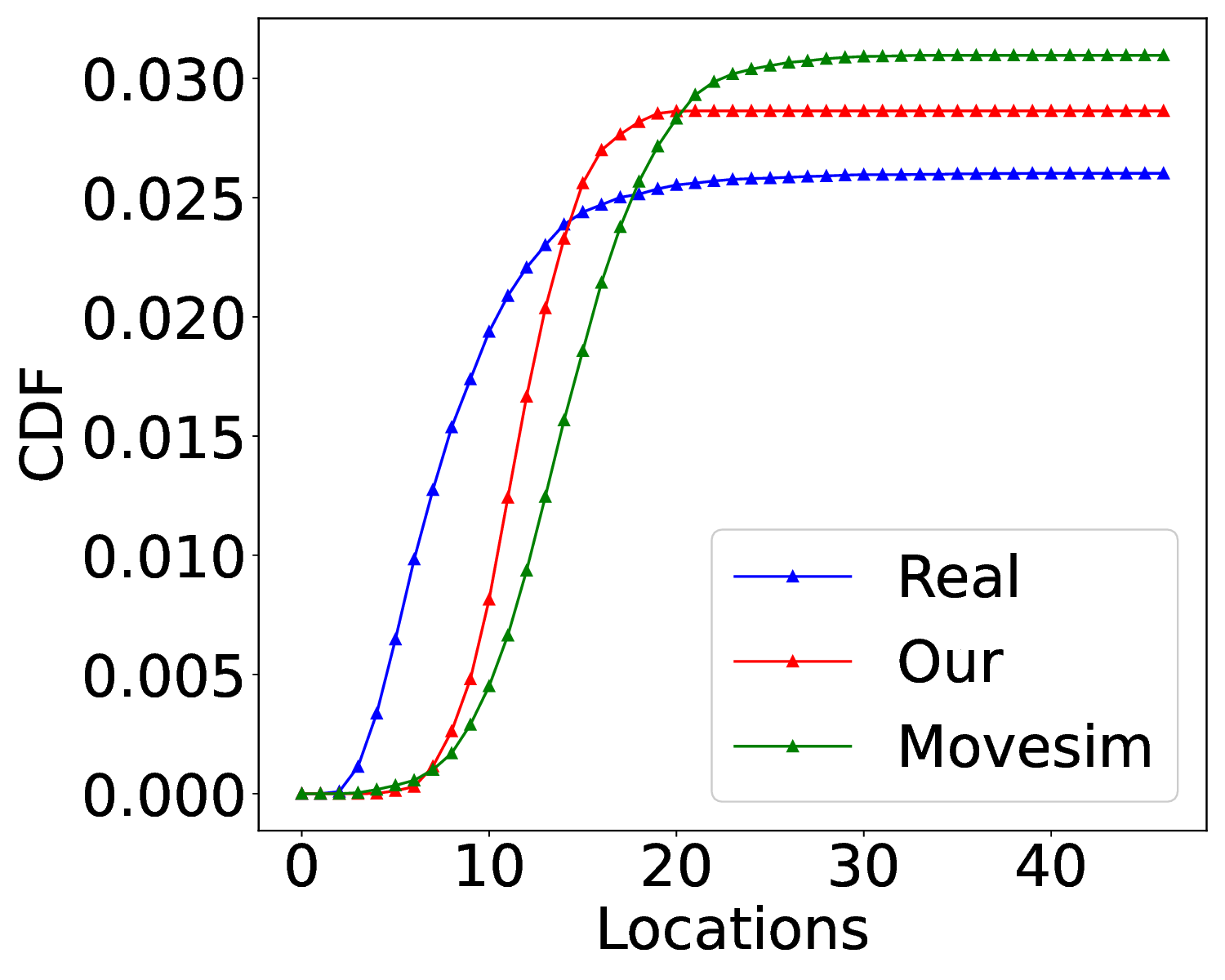}}
\subfigure[]{\includegraphics[width=.22\textwidth]{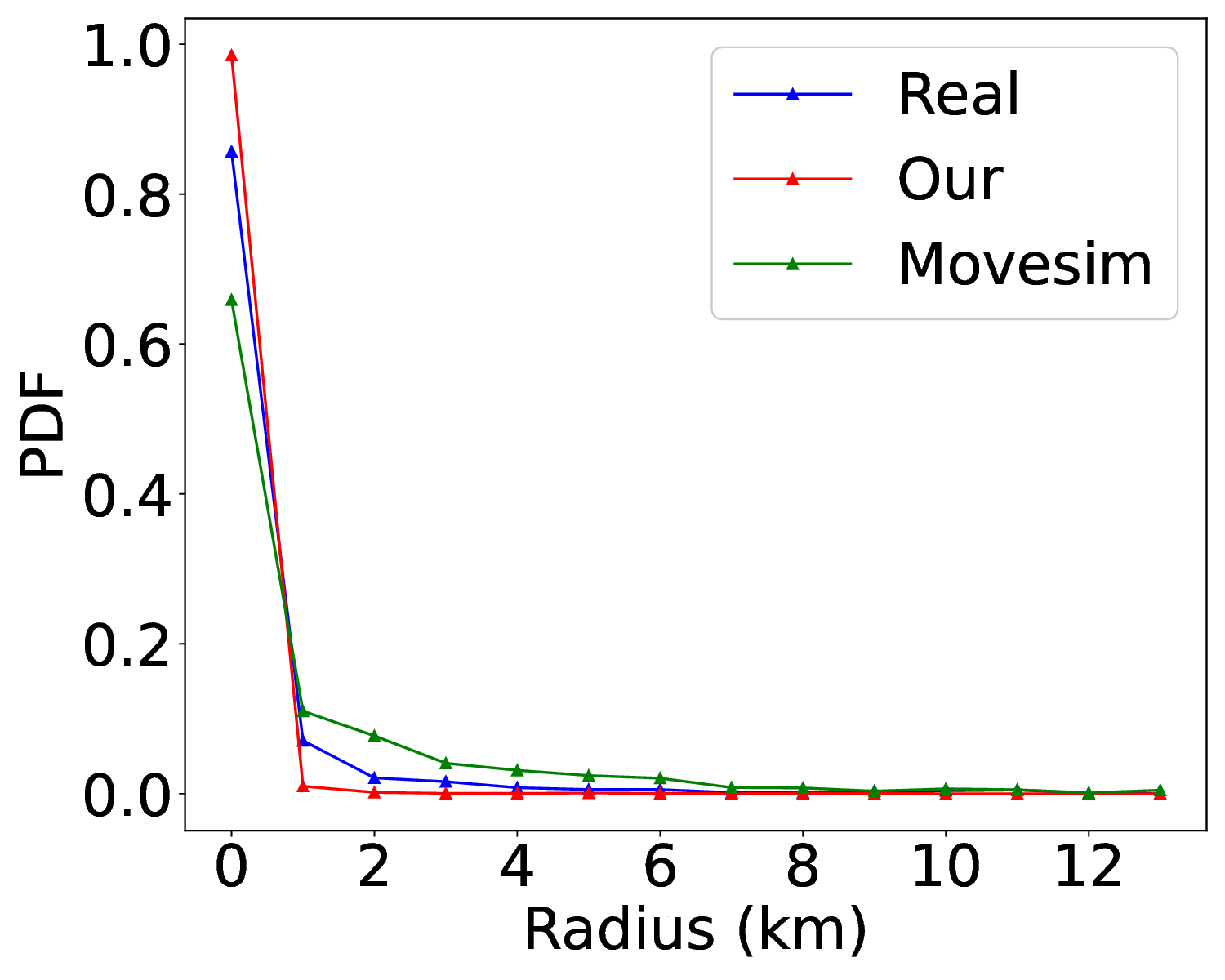}}
\subfigure[]{\includegraphics[width=.22\textwidth]{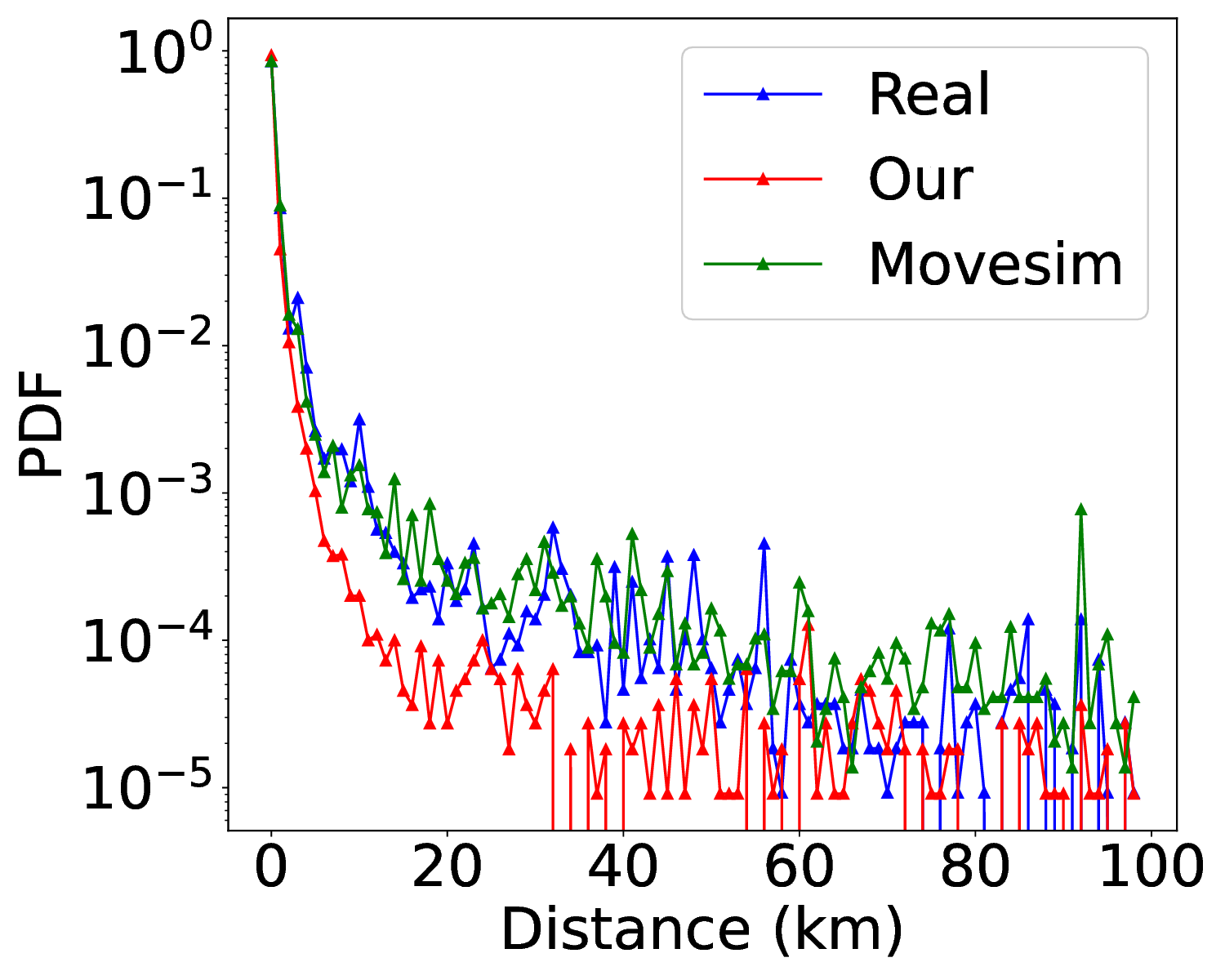}}
\subfigure[]{\includegraphics[width=.22\textwidth]{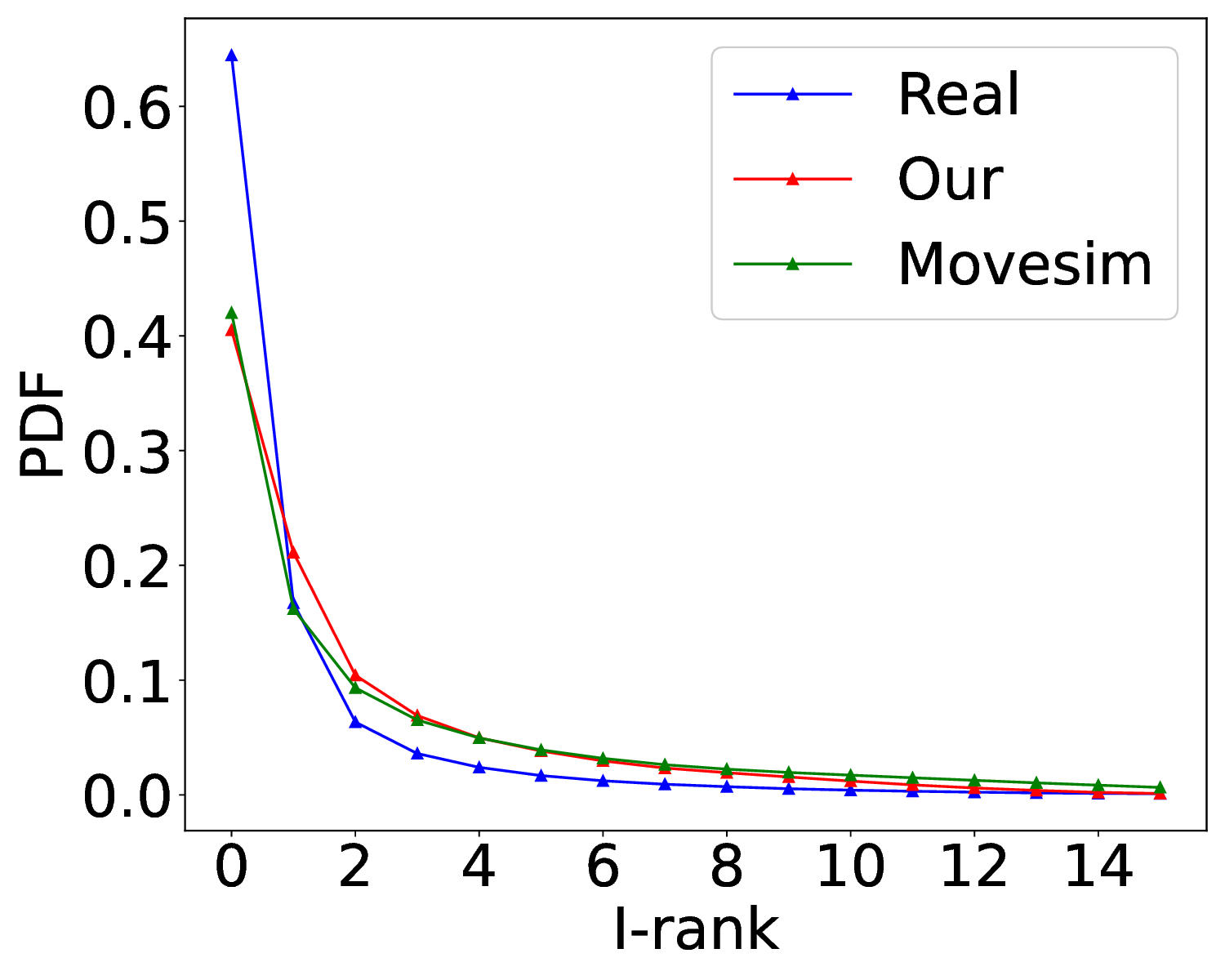}}
\vspace{-0.4cm}
\caption{Visualization of the distribution of the selected statistical metrics on the GeoLife dataset.} \label{fig:dis_geolife}
\vspace{-0.2cm}
\end{figure}

\begin{table}[t!]
\begin{center}
\caption{A list of commonly used notations.}
\begin{tabu}{|p{2cm}<{\centering}|[1.02pt]p{6.0cm}|}
\tabucline[1.02pt]{-}
\textbf{Notation} & \textbf{Description} \\ \hline
 $\mathcal{U}$ & The set of users.\\
 $\mathcal{L}$ & The set of geographical regions.\\
 $T_u$ & The user trajectory for user $u\in\mathcal{U}$.\\
 $\mathcal{S}$ & The state space.\\
 $\mathcal{A}$ & The action space.\\
 $P(s_{t+1}|s_t,a_t)$ & The state transition probability.\\
 $\pi_\theta(a|s)$ & The policy function.\\
 $D^u_{\phi_u}(s,a)$ & The personal discriminator of user $u\in\mathcal{U}$.\\
 $R(s,a)$ & The reward function obtained from the private aggregation mechanism.\\
 $\xi(s,a)$ & The variation of the rewards across different users.\\
 $\hat{R}(s,a)$ & The reward function combining the private aggregation mechanism and the reward dynamics.\\
 $\beta$ & Hyper-parameter to adjust the influence of the reward dynamics compensation mechanism.\\
 ${\rm Laplace}(a,b)$ & Laplace noise with the mean $a$ and the scale $b$.\\
 $\lambda$ & The scale of the Laplace noise added to the private aggregation mechanism.\\
 $\lambda_c$ & The scale of the Laplace noise added to the reward dynamics compensation mechanism.\\
\tabucline[1.02pt]{-}
\end{tabu}
\label{tab:symbal}
\end{center}
\vspace{-0.4cm}
\end{table}

\renewcommand{\algorithmicrequire}{\textbf{Input:}}
\begin{algorithm*}[t!]
\caption{PateGail}\label{alg:framework}
\begin{algorithmic}[1]
\REQUIRE
Expert trajectories $\{T_u\}_{u\in\mathcal{U}}$.
\STATE
Initialize $\pi_{\theta}$, $\{D_{\phi_u}\}_{u\in\mathcal{U}}$ with random weights;
\FOR{$i$=0,1,2...}
\STATE
\underline{\textbf{Server implements steps 4-5:}}

\STATE
Generate a batch of trajectory $T_i$$\sim$$\pi_\theta$;
\STATE
Send $T^G_i$ to each user device $u\in\mathcal{U}$;

\STATE
\underline{\textbf{Clients implement steps 7-12 for $u\in\mathcal{U}$ in parallel:}}
\FOR{$k=0,1,2...$}
\STATE
Calculate a batch of rewards $D_{\phi_u}(s,a)$ for each state-action pair in trajectories $T^G_i$;
\STATE
Sample synthetic trajectories $T^G_{ki}$ based on $\pi_\theta$;
\STATE
Sample expert trajectories $T^E_k$ from the local trajectory data $T_u$ with the same batch size;
\STATE
Update $D_{\phi_u}$ based on (\ref{equ:teacherdisc}) Adam Optimizer with the positive samples $T^E_k$ and negative samples $T^G_{ki}$.
\ENDFOR
\STATE
\underline{\textbf{Server implements steps 14-15:}}
\STATE
Calculate the aggregated reward $\hat{R}(s,a)$ based on (\ref{equ:rlowerbound});
\STATE
Update $\pi_\theta$ with reward $\hat{R}(s,a)$ via PPO method;
\ENDFOR
\end{algorithmic}
\end{algorithm*}

\end{document}